\documentclass{article}
\usepackage{log_2023}						

\usepackage{booktabs}						
\usepackage{multirow}						
\usepackage{amsfonts}						
\usepackage{graphicx}						
\usepackage{duckuments}						

\usepackage[numbers,compress,sort]{natbib}	

\usepackage{wrapfig}
\usepackage[utf8]{inputenc}
\usepackage{verbatim}
\newsavebox{\mintedbox}
\usepackage{xcolor}
\usepackage{float}
\usepackage{algorithmic}
\usepackage[draft, newfloat]{minted}
\usepackage{mathtools}
\usepackage{setspace}
\usepackage{placeins}
\usepackage{tcolorbox}

\usepackage{listings}
\usepackage{caption}

\usepackage{pgfplots}

\usepackage[justification=centering]{subcaption}

\newcommand{\bb}{\hspace{-1mm} $\bullet$}

\newenvironment{code}{\captionsetup{type=listing}}{}

\usepackage{pifont}

\definecolor{cycle2}{RGB}{106, 191, 0}
\definecolor{cycle3}{RGB}{191, 0, 0}

\newcommand{\cmark}{\textcolor{cycle2}{\ding{52}}} %
\newcommand{\xmark}{\textcolor{cycle3}{\ding{56}}}

\newcommand{\ER}{Erd\H{o}s-R\'enyi}   

\newcommand{\specialcell}[2][c]{%
  \begin{tabular}[#1]{@{}c@{}}#2\end{tabular}}

\definecolor{cycle2}{RGB}{106, 191, 0}
\definecolor{cycle3}{RGB}{191, 0, 0}

\usepackage{bbm}

\title[PyTorch Geometric Signed Directed]{PyTorch Geometric Signed Directed: A Software Package on Graph Neural Networks for Signed and Directed Graphs}

\author[Y. He et al.]{Yixuan He\\
University of Oxford\\
\email{yixuan.he@balliol.ox.ac.uk}\And
Xitong Zhang\\
Michigan State University\\
\email{zhangxit@msu.edu} \And
Junjie Huang\\
Southwest University\\
\email{junjiehuang@swu.edu.cn}\And
Benedek Rozemberczki\\
Isomorphic Labs\\
\email{benedek.rozemberczki@gmail.com}\And
Mihai Cucuringu \& Gesine Reinert\\
Univerisity of Oxford \& The Alan Turing Institute\\
\email{\{mihai.cucuringu, reinert\}@stats.ox.ac.uk}
}

\begin{document}

\maketitle

\begin{abstract}
Networks are ubiquitous in many real-world applications (e.g., social networks encoding trust/distrust relationships, correlation networks arising from time series data). While many networks are signed or directed, or both, there is a lack of unified software packages on graph neural networks (GNNs) specially designed for signed and directed networks. In this paper, we present PyTorch Geometric Signed Directed (PyGSD), a software package which fills this gap. Along the way, we evaluate the implemented methods with experiments with a view to providing insights into which method to choose for a given task. The deep learning framework consists of easy-to-use GNN models, synthetic and real-world data, as well as task-specific evaluation metrics and loss functions for signed and directed networks. As an extension library for PyG, our proposed software is maintained with open-source releases, detailed documentation, continuous integration, unit tests and code coverage checks. The \textit{GitHub} repository of the library is \url{https://github.com/SherylHYX/pytorch_geometric_signed_directed}.
\end{abstract}

\vspace{-5mm}
\section{Introduction}
\vspace{-2mm}
Graph data arise in many areas, such as social networks, citation networks, and biochemical graphs. Such data are related to many learning problems \cite{Wu, Zhou}. 
To tackle network inference tasks, graph neural networks (GNNs) are a useful tool.  Based on deep learning techniques such as network architecture, optimization approaches and parallel computation, GNNs can be trained just like standard neural networks. By leveraging the network structure, GNNs are able to retain information from the  nodes' neighborhood  and incorporate long-range dependencies \cite{Zhou}.
GNNs have a wide range of applications, such as social network analysis \cite{he2022sssnet}, traffic predictions \cite{liu2021traffic}, biological applications \cite{li2021graph}, recommender systems, computer vision, and natural language processing \cite{Wu}.

\looseness=-1 Although traditional network analysis usually focuses on a single fixed simple network, which could often be represented by a symmetric adjacency matrix with nonnegative entries, intricate network types are often more realistic. 
Signed networks, which have positive or negative edge weights, have been of interest in social network analysis \cite{huang2021sdgnn, sampson1969novitiate}, 
with signs indicating positive or negative sentiments. 
The development of signed network algorithms has been motivated for example by the task of clustering time series with regards to correlations~\cite{aghabozorgi2015time}, including economic time series that capture macroeconomic variables \cite{focardi2001clustering}, and financial applications \cite{ziegler2010visual, pavlidis2006financial}. 
The empirical correlation matrix can be constructed as a weighted signed network, with signs indicating correlations, potentially sparsified after thresholding. Thresholds can be obtained for example 
as p-values of statistical tests of significance for individual correlations. Many signed networks are also directed, in that they have asymmetric sending and receiving patterns. 
Indeed, edge directionality could play an integral role in processing and learning from network data \cite{marques2020signal}; directed networks, even unsigned,  have many applications such as clustering time-series data with lead-lag relationships 
\cite{bennett2021detection}, detecting influential groups in social networks \cite{bovet2020activity, he2022digrac, zhang2021magnet}, ranking \cite{he2022gnnrank}, angular synchronization~\cite{he2023robust}, and biological applications \cite{marques2020signal}. 

We note that there have been several GNNs~\cite{schlichtkrull2018modeling, vashishth2019composition, zhang2020relational} introduced for multi-relational graphs, i.e., graphs with different types of edges. There is typically a linear relationship between the number of parameters that can be learned in such networks and the number of edge types. In general, weighted signed graphs with real-valued edge weights are not just special cases of multi-relational graphs. If the graph is unweighted or the weighting function takes a finite number of values, however, we could view signed graphs as special multi-relational graphs.  
Nevertheless, there is an implicit relationship between the different edge types in signed graphs (possibly weighted), namely that edges with large weights are considered more important than edges with small weights and that negative edges are interpreted as the opposite of positive edges. As a result, if we consider an arbitrary multi-relational graph, we would have much more trainable parameters than using these relationships. Instead, GNN methods which are tailored for signed and directed networks, as presented in this library, are a useful addition to the tool set of GNNs for network analysis. 


\looseness=-1 Moreover, deep learning frameworks \cite{pytorch, abadi2016tensorflow, mxnet, jax} have facilitated the emergence of open-source software on deep learning on graphs \cite{pytorch_geometric, rozemberczki2021pytorch, dlg, hu2021efficient, spektral, StellarGraph}. Such open-source libraries are essential to the practical success and large-scale deployment of graph machine learning systems \cite{rozemberczki2021pytorch}.  However, despite the popularity of signed and directed networks, there does not exist an open-source library to include GNNs as well as related data processing methods on them. This motivates the design of our open-source library, called \textit{PyTorch Geometric Signed Directed} (PyGSD), which is an extension library of the popular \textit{PyTorch Geometric} (PyG) \cite{pytorch_geometric}, but is specially designed for signed and directed networks.

\textbf{Main contributions.} \bb (1) We present \textit{PyTorch Geometric Signed Directed} (PyGSD), the first deep learning library for graph neural networks in signed and directed networks. This is an open-source library equipped with public releases, documentation, code examples, continuous integration, unit tests and code coverage checks. \bb (2)  We provide easy-to-use GNN models, synthetic data generators and real-world data loaders, as well as task-specific evaluation metrics and loss functions for signed and directed networks in PyGSD. \bb (3)  We evaluate task performance of GNNs available in PyGSD and provide insights into which methods to choose for which tasks.

\vspace{-4mm}
\section{Background}
\vspace{-2mm}
\label{sec:theo}
In this paper, let $\mathcal{G}=(\mathcal{V}, \mathcal{E}, w, \mathbf{X}_\mathcal{V})$ denote
a (possibly signed, directed and weighted) network with node attributes, 
where $\mathcal{V}$ is the set of nodes, $\mathcal{E}$ is the set of (directed) edges or links, and $w \in (-\infty, \infty)^{\lvert \mathcal{E} \rvert}$ is the set of weights of the edges. The network $\mathcal{G}$ has adjacency matrix $\mathbf{A} = (A_{ij})_{i,j \in \mathcal{V}} $, 
where $\mathbf{A}_{ij}=w_{ij}$, the edge weight, if there is an edge from node $v_i$ to node $v_j$; 
otherwise $\mathbf{A}_{ij}=0$.  
Here, $\mathcal{G}$ could have self-loops, but no multiple edges.
The total number of nodes is $n=\lvert \mathcal{V} \rvert$, and
$\mathbf{X}_\mathcal{V}\in \mathbb{R}^{n\times d_\text{in}}$ is an attribute matrix whose rows are node attributes (which could be generated from $\mathbf{A}$), where $d_\text{in}$ denotes the input attribute dimension.
For a signed network, we split the edge set into positive and negative subsets $\mathcal{E}=\mathcal{E}^+\cup\mathcal{E}^-,$
where the weights for the edges in $\mathcal{E}^+$ and $\mathcal{E}^-$ are positive and negative, respectively. Here we use {\it network} and {\it graph} interchangeably.
\vspace{-3mm}
\subsection{Insights from Social Network Analysis: Social Balance Theory and Status Theory}
\vspace{-2mm}
\looseness=-1 Signed and possibly directed 
social networks are addressed by structural balance theory~\cite{heider1946attitudes} and status theory~\cite{leskovec2010signed} from sociology.
A network is called {\it balanced}  if all cycles have an even number of negative edges; otherwise it is unbalanced. 
Structural balance theory, proposed by Heider in 1946~\cite{heider1946attitudes}, hypothesizes how humans are motivated to change their attitudes when their social networks are unbalanced. Its most famous example is  the principle that \textit{the friend of my friend is my friend and the  enemy of my enemy is my friend}. Unlike balance theory, which is usually applied to undirected signed networks, status theory~\cite{guha2004propagation} considers the direction in signed directed networks.
It supposes that a positive directed link ``+'' from A to B indicates that \textit{target node B has higher status than source node A}; a negative directed link ``-'' from A to B indicates that \textit{source node A has higher status than target node B}.

Incorporating such insights from social network analysis into GNNs requires specific tailoring and hence the development of particular tools. In particular, both sociological theories can be associated with signed triangles~\cite{chen2018bridge, huang2021sdgnn}. Related works on graph representation learning which model the sociological theory described above include for example SGCN~\cite{Derr}, SiGAT~\cite{huang2019signed}, SNEA~\cite{li2020learning}, and SDGNN~\cite{huang2021sdgnn}.
In PyGSD, these models are implemented, together with their loss functions (e.g., Structural Balance Loss in SGCN~\cite{Derr} and Signed Direction Loss in SDGNN~\cite{huang2021sdgnn}).
\vspace{-3mm}
\subsection{Spatial Graph Neural Networks}
\vspace{-2mm}
Spatial GNNs utilize information propagation among nodes to define convolutions on graphs~\cite{Wu}. 
The Message Passing Neural Network (MPNN) by~\cite{gilmer2017neural} outlines a general framework of spatial GNNs, treating graph convolutions as a message-passing procedure among nodes and edges. The general form of message passing is defined as $\mathbf{h}_i^{(l)}=\text{UPDATE}^{(l-1)}\left(\mathbf{h}_i^{(l-1)},\text{AGGREGATE}^{(l-1)}(\{\mathbf{h}_j: \mathbf{A}_{i,j}\neq 0\})\right)$ where $\text{UPDATE}$ and $\text{AGGREGATE}$ are arbitrary differentiable functions. 
However, these spatial methods naturally  are not tailored to the analysis of signed and directed networks. In particular they do not take the specific insights from social network analysis into account, but can be adapted to do so.
Some examples of such spatial GNNs for signed/directed graphs that could be viewed as instantiations of MPNNs are \cite{he2022digrac} and \cite{Derr}. For signed networks, the aggregation function $\text{AGGREGATE}$ provides the option to consider positive and negative neighbors separately, such as in \cite{Derr} and \cite{he2022sssnet}. The attention mechanism, which assigns an attention weight or importance to each neighbor, is also considered in signed directed GNNs under the MPNN framework~\cite{li2020learning, huang2019signed, huang2021sdgnn}. Indeed, spectral GNNs could be implemented in a spatial way, as a special form of MPNN, as validated by \cite{zhang2021magnet} and \cite{he2022msgnn}. 

\vspace{-3mm}
\subsection{Some Tools from Spectral Analysis: Graph Convolution and Laplacian Matrices}
\vspace{-2mm}
\label{sec:spectral_theory}
For a graph signal $\mathbf{x} \in \mathbb{R}^{\lvert \mathcal{V} \rvert}$ on an undirected graph $\mathcal{G}$, the conventional graph convolution is defined as
$g_{\theta} \star \mathbf{x}=\mathbf{U} g_{\theta} \mathbf{U}^{\top} \mathbf{x}$, where $g_{\mathbf{\theta}}=\mathrm{diag}(\mathbf{\theta}), \mathbf{\theta} \in \mathbb{R}^{{\lvert \mathcal{V} \rvert}}$,
and $\mathbf{U}$ is obtained from the eigendecomposition of the normalized Laplacian matrix: $\mathbf{L}=\mathbf{I}_{\lvert \mathcal{V} \rvert}-\mathbf{D}^{-\frac{1}{2}} \mathbf{A} \mathbf{D}^{-\frac{1}{2}} = \mathbf{U}\Lambda\mathbf{U}^\top,$ where $\mathbf{D}_{ii}=\sum_j \mathbf{A}_{ij}$ and $\mathbf{U}^\top \mathbf{x}$ is the graph Fourier transform. \cite{Defferrard2016} and \cite{kipf2016semi} proposed related convolutional networks based on $\mathbf{L}$, but both of them theoretically require a symmetric adjacency matrix to have a complete set of eigenvectors, which is not usually satisfied for directed graphs, whose edges often have asymmetric sending and receiving patterns. \cite{tong2020directed} and \cite{tong2020digraph} proposed symmetric matrices to incorporate the direction. \cite{zhang2021magnet} uses a complex-valued Laplacian matrix, encoding direction in the phase matrix and weights in the magnitude matrix.

\vspace{-5mm}
\section{Tasks and Evaluation Metrics} \vspace{-2mm}
\label{sec:tasks} 
In this section, we review the typical problem setup, discuss typical tasks and loss functions as well as evaluation metrics, for signed and directed graphs. Typical machine learning tasks in signed/directed graphs include: 1) link prediction, 2) node classification, 3) node clustering. These tasks can be carried out in a supervised (a large proportion of the data are training data with label supervision), semi-supervised (label supervision on a small proportion of data), or unsupervised (no label supervision) setting. Here we do not include tasks relating to groups of graphs or to synthetic graph generation. 

\vspace{-3mm}
\subsection{Link Prediction}
\vspace{-2mm}
\label{subsec:link_pred}
For \emph{signed} networks, a typical task is to predict the sign of an existing edge, i.e., link sign prediction (SP). One typical GNN pipeline first learns node embeddings based on some loss function dependent on, for example,  node embeddings and edge signs \cite{Derr, huang2021sdgnn}, and then applies a binary classifier such as the logistic regression classifier to output the final predictions. Another GNN pipeline conducts end-to-end training using a differentiable loss function such as the binary cross entropy loss~\cite{he2022msgnn}.

For \emph{directed} networks, the three typical link prediction tasks are: 1)~Direction prediction (DP): predict the edge direction of pairs of vertices $v_i, v_j$ for which either $(v_i, v_j) \in \mathcal{E}$ or $(v_j, v_i) \in \mathcal{E}$. 
2)~Existence prediction: predict if $(v_i, v_j) \in \mathcal{E}$ by considering ordered pairs of vertices $(v_i, v_j)$. 
3)~Three-class classification (3C): classify an edge $(v_i, v_j) \in \mathcal{E}$, $(v_j, v_i) \in \mathcal{E}$, or $(v_i, v_j), (v_j, v_i) \notin \mathcal{E}$. 

The link prediction tasks mentioned above could be conducted on signed and directed networks, but the task focus is on either link sign or directionality but not both. In this library, we also provide more general tasks with edge splitting tools related to both link signs and directionality: 4) Four-class classification (4C): classify an edge $(v_i, v_j) \in \mathcal{E}^+$, $(v_j, v_i) \in \mathcal{E}^+$, $(v_i, v_j) \in \mathcal{E}^-$, or $(v_j, v_i) \in \mathcal{E}^-$; 5) Five-class classification (5C): in addition to the classes in 4), add a class $(v_i, v_j), (v_j, v_i) \notin \mathcal{E}$.

\looseness=-1 For splitting edges into training/test sets, for training and evaluation we discard edges of the input network that fall into more than one classification category, but we keep these edges for the GNN during training. 
When treated as a classification problem, the cross-entropy loss function can be employed \cite{zhang2021magnet}. Results are typically evaluated with accuracy. The  Area Under the Receiver Operating Characteristic Curve (AUC) \cite{bradley1997use} and the  F1 score \cite{sokolova2009systematic} are also employed in binary classification scenarios.

\vspace{-3mm}
\subsection{Node Classification}
\vspace{-2mm}
\looseness=-1 Node classification classifies each node in a graph into one of $K$ classes. The cross-entropy loss function is usually applied for training, and performance is often evaluated via accuracy. The training setup is usually either fully-supervised, with a high proportion of training nodes, or semi-supervised, where only a small proportion of nodes in each class are provided with label information during training.
\vspace{-3mm}
\subsection{Node Clustering}
\vspace{-2mm}
A clustering into $K$ clusters is 
a partition of the node set 
into disjoint sets
$\mathcal{V}=\mathcal{C}_0\cup\mathcal{C}_1\cup\cdots\cup\mathcal{C}_{K-1}.$ 
Intuitively, nodes within a cluster should be similar to each other, while nodes across clusters should be dissimilar. 
In a semi-supervised setting, for each of the $K$ clusters, a fraction of training nodes are selected as seed nodes, for which 
the cluster membership labels are known before training. 
The set of seed nodes is denoted as $\mathcal{V}^\text{seed}\subseteq\mathcal{V}^\text{train}\subset\mathcal{V},$
where $\mathcal{V}^\text{train}$ is the set of all training nodes.
For this task, 
the goal is to use the embedding for assigning each node $v_i \in \mathcal{V}$ to a cluster containing known seed nodes. 
When no seeds are given, we are in a self-supervised setting, where only the number of clusters, $K,$ is given. The quality of a clustering partition is often assessed through a modularity objective function \cite{traag2019louvain} comparing the partition to that expected under a null model for the network, with the assumption that nodes within a cluster are relatively more densely connected than nodes across clusters. However, depending on the task at hand, \emph{similarity} could have different meanings \cite{he2022gnns}.
In a signed network with positive and negative edges, similarity may relate to the neighorhood of a node such as the proportion of shared friends or enemies.
In a directed network, nodes could also be clustered with regard to their position within a directed flow on the network,
see \cite{he2022digrac}.

\begin{wraptable}{r}{0.54\linewidth}
\vspace{-5mm}
\caption{Summary statistics 
for the real-world networks that can be loaded by \textit{PyTorch Geometric Signed Directed }(PyGSD). Here $n$ is the number of nodes, $\lvert\mathcal{E}^+\rvert$ and $\lvert\mathcal{E}^-\rvert$ denote the number of positive and negative edges, respectively. For an unsigned network, $\lvert\mathcal{E}^-\rvert=0.$ For an unweighted graph $\lvert\mathcal{E}^-\rvert=0$ means that the set of edge weights only contains one value if we disregard signs. ``Labeled" denotes whether node labels are provided. Note that for labeled data sets, all nodes are labeled. We can conduct link prediction and node clustering tasks on all data sets, while node classification is only possible on labeled data sets. In the last two columns we report whether the network is directed or weighted. Note that statistics for Fin-YNet, FiLL-pvCLCL, FiLL-OPCL, and Lead-Lag are averaged over 21, 21, 21 and 19 financial networks, respectively.}
\vspace{-1mm}
\centering
\setlength\tabcolsep{2pt}
\resizebox{\linewidth}{!}{\begin{tabular}[htp!]{lrrrrrr}
\toprule
Data set               & $n$ &  $\lvert\mathcal{E}^+\rvert$ & $\lvert\mathcal{E}^-\rvert$&Labeled & Directed &Weighted\\ \midrule
Sampson                 & 25&148&182&\cmark&\xmark&\cmark \\ 
Cornell&183&298&0&\cmark&\cmark&\xmark\\
Texas&183&325&0&\cmark&\cmark&\xmark\\
Telegram&245&8,912&0&\cmark&\cmark&\cmark\\
Wisconsin&251&515&0&\cmark&\cmark&\xmark\\
Lead-Lag&269&29,159&0&\xmark&\cmark&\cmark\\
Rainfall&306&64,408&29,228&\xmark&\xmark&\cmark\\
FiLL-OPCL&430&84,467&100,013&\xmark&\cmark&\cmark\\
FiLL-pvCLCL&444&84,677&112,015&\xmark&\cmark&\cmark\\
Fin-YNet &451&148,527&54,313&\xmark&\xmark&\cmark\\
S\&P 1500                &1,193&1,069,319&353,930&\cmark&\xmark&\cmark      \\ 
Blog&1,222&19,024&0&\xmark&\cmark&\cmark\\
Chameleon&2,277&36,101&0&\cmark&\cmark&\xmark\\
Cora-ML&2,995&8,416&0&\cmark&\cmark&\xmark\\
PPI&3,058&7,996&3,864&\xmark&\xmark&\cmark  \\ 
Migration&3,075&721,432&0&\xmark&\cmark&\cmark\\
CiteSeer&3,312&4,715&0&\cmark&\cmark&\xmark\\
BitCoin-Alpha & 3,783 & 22,650 & 1,536 &\xmark& \cmark&\cmark \\
Squirrel&5,201&222,134&0&\cmark&\cmark&\cmark\\
BitCoin-OTC & 5,881 &32,029  & 3,563&\xmark & \cmark&\cmark\\
Wiki-Rfa              & 7,634&135,753&37,579&\xmark&\cmark&\cmark      \\
WikiCS&11,701&297,110&0&\cmark&\cmark&\xmark\\
Slashdot & 82,140 &380,933& 119,548     &\xmark& \cmark  &\xmark\\
Epinions & 131,580& 589,888 &121,322     &\xmark& \cmark &\xmark \\
WikiTalk&2,388,953&5,018,445&0&\xmark&\cmark&\xmark\\
\bottomrule
\end{tabular}}
\label{tab:loaded_data_sets}
\vspace{-8mm}
\end{wraptable}
\looseness=-1 A GNN's objective plays an essential role in guiding the GNN to learn. In a node clustering task, when seed nodes with known cluster labels are available, the cross entropy loss function is usually applied. In \cite{he2022sssnet}, a contrastive loss function based on triplets of the nodes is used to push embeddings of nodes within clusters to be closer to each other than  to nodes in other clusters. 
For unsupervised tasks, however, objectives which are independent of known labels are required. Differentiable objectives can then be defined based on the tasks at hand. A probabilistic version of the balanced normalized cut \cite{Chiang} loss is proposed in \cite{he2022sssnet}, while a probabilistic version of flow imbalance loss is introduced in \cite{he2022digrac}. 

For a node clustering problem, as cluster indices could be permuted, accuracy would not be an appropriate evaluation measure. Instead, the Adjusted Rand Index (ARI) \cite{hubert1985comparing} is often chosen, which is invariant to label permutation. Depending on the downstream task, other measures could be used, such as the \emph{unhappy ratio} by \cite{he2022sssnet}, and the \emph{imbalance scores} by \cite{he2022digrac}. 
\vspace{-5mm}
\section{Data Sets} \label{sec:data} 
\vspace{-2mm}
\textbf{Synthetic Data.}
\label{subsec:synthetic_data}
PyGSD provides synthetic data generators for Signed Stochastic Block Models(SSBMs) and Polarized SSBMs from \cite{he2022sssnet}, Directed Stochastic Block Models (DSBMs) from \cite{he2022digrac}, as well as Signed Directed Stochastic Block Models (SDSBMs) from \cite{he2022msgnn}, which are typical synthetic signed and directed networks used in related research papers. Details of these synthetic models can be found in Appendix~\ref{appendix_sec:synthetic_description}.

\textbf{Real-world Data Sets.}
Table~\ref{tab:loaded_data_sets} summarizes real-world data sets for which PyGSD provides data loaders. These are real-world data sets used in related research papers, with network size ranging from 25 to 2,388,953 nodes. Details of these data sets are given in Appendix~\ref{appendix_sec:real_world_description}. 
\vspace{-4mm}
\section{Methods Currently in PyGSD}
\label{sec:methods}
\vspace{-1mm}
\looseness=-1 In this section, we 
review GNNs which are currently implemented in PyGSD and which will be used in our experiments in Sec.~\ref{experiments}. GNNs can be broadly classified into spatial or spectral, depending on whether or not they utilize the spectral theory outlined in Sec.~\ref{sec:spectral_theory}. Table~\ref{tab:methods} lists whether the implemented GNNs can deal with signed edges or directed edges, together with the tasks they are concerned with, and whether the method is spatial or spectral. Note that non-GNN baselines are not implemented in our library, but we refer readers to \cite{he2022sssnet} and \cite{he2022digrac} for their implementations. 

\textbf{Directed Unsigned Networks.} 
Research on directed network analysis has been mainly spectral in nature, and centered around symmetrization-based techniques \cite{satuluri2011symmetrizations, ma2019spectral, columbia_directed}, but edge directionality itself can contain important information \cite{marques2020signal, cucuringu2020hermitian}. Imbalanced flows in directed networks have been uncovered via Hermitian clustering \cite{cucuringu2020hermitian} and motif-based techniques \cite{MotifSpectralDirectedClustering_ANS_2020}. Recently, researchers have started applying GNNs to extract essential information from directed edges.

DGCN \cite{tong2020directed} uses first and second order proximity, and constructs three Laplacians, but can be space and speed-inefficient.
DiGCN \cite{tong2020digraph} simplifies DGCN, builds a directed Laplacian based on PageRank, and aggregates information dependent on higher-order proximity. We denote the variant with the so-called ``inception blocks" as DiGCNIB, and the variant without ``inception blocks" as DiGCN.  
MagNet \cite{zhang2021magnet} constructs a Hermitian matrix that encodes undirected geometric structure in the magnitude of its entries, and directional information in their phase. The Laplacian matrices proposed by \cite{tong2020directed,tong2020digraph} and \cite{zhang2021magnet} are based on the spectral theory sketched  
in Sec.~\ref{sec:spectral_theory}.  
DiGCL \cite{tong2021directed} introduces a directed network data augmentation method called \textit{Laplacian perturbation} and conducts directed network contrastive learning.  
DIGRAC \cite{he2022digrac} conducts directed network clustering based on flow imbalance measures, with novel imbalance objectives and evaluation metrics.

\begin{wraptable}{r}{0.6\textwidth}
\centering
\vspace{-4mm}
\caption{
Signed/directed GNNs implemented in PyGSD}\label{tab:methods}
\vspace{-1mm}
 \resizebox{\linewidth}{!}{
\begin{tabular}{lcccccc}
\toprule
    Model    & \specialcell{Signed\\ Edges}   & \specialcell{Directed\\ Edges}    & \specialcell{Node-Level\\ Tasks}    & \specialcell{Edge-Level\\ Tasks}&\specialcell{Spatial or\\Spectral} \\ \midrule
DGCN \cite{tong2020directed}&\xmark&\cmark&\cmark&\cmark&Spectral\\
DiGCN \cite{tong2020digraph}&\xmark&\cmark&\cmark&\cmark&Spectral\\
DiGCNIB \cite{tong2020digraph}&\xmark&\cmark&\cmark&\cmark&Spectral\\
MagNet \cite{zhang2021magnet}&\xmark&\cmark&\cmark&\cmark&Spectral\\
DiGCL \cite{tong2021directed}&\xmark&\cmark&\cmark&\xmark&Spectral\\
DIGRAC \cite{he2022digrac}&\xmark&\cmark&\cmark&\xmark&Spatial\\
SGCN \cite{Derr}&\cmark&\xmark&\xmark&\cmark&Spatial\\
SiGAT \cite{huang2019signed}&\cmark&\cmark&\xmark&\cmark&Spatial\\
SNEA \cite{li2020learning}&\cmark&\cmark&\xmark&\cmark&Spatial\\
SDGNN \cite{huang2021sdgnn}&\cmark&\cmark&\xmark&\cmark&Spatial\\
SSSNET \cite{he2022sssnet}&\cmark&\cmark&\cmark&\xmark&Spatial\\
MSGNN \cite{he2022msgnn}&\cmark&\cmark&\cmark&\cmark&Spectral\\
\bottomrule
\end{tabular}
}
\vspace{-5mm}
\end{wraptable}
\textbf{Signed Undirected and Signed Directed Networks.}
Within the last decade, the landscape of signed network analysis has been dominated by non-GNN methods (in particular spectral methods), such as those based on the (potentially normalized) Signed Laplacian matrix \cite{kunegis2010spectral} and its variations \cite{zheng2015spectral, mercado2016clustering}, Balanced Normalized Cut and Balanced Ratio Cut \cite{Chiang}, and many others \cite{Yuan2017, wang2017attributed,SPONGE_AISTATS_2019}. 

\looseness=-1 More recently, GNNs for signed networks emerged.
SGCN \cite{Derr} proposes an information aggregation and propagation mechanism with the mean-pooling strategy for signed undirected networks based on social balance theory \cite{harary1953notion}. 
SiGAT \cite{huang2019signed} utilizes graph attention network GAT~\cite{velivckovic2017graph} in embedding learning for signed directed networks and devises a motif-based GNN architecture based on balance theory and status theory. SNEA \cite{li2020learning} proposes another graph attentional layer, which uses a masked self-attention mechanism, and designs an objective function for both framework optimization and node representation learning. SDGNN \cite{huang2021sdgnn}, though also using graph attention, is more efficient than SiGAT, and its objective functions can model not only the edge sign, but in addition other vital features such as edge directions and triangles. 
SSSNET \cite{he2022sssnet} conducts the semi-supervised node clustering task in signed networks, with a novel aggregation scheme that is not based on the popular social balance theory \cite{harary1953notion} that many previous signed GNNs rely on.
The very recent  MSGNN~\cite{he2022msgnn} introduces a novel magnetic signed Laplacian as a natural generalization of both the signed Laplacian on signed graphs and the magnetic Laplacian on directed graphs, and proposes a spectral GNN via this magnetic Laplacian matrix, extending the ideas from Sec.~\ref{sec:spectral_theory} to signed directed networks.

\vspace{-4mm}
\section{Framework Design of PyGSD}\label{sec:framework} 
\vspace{-1mm}
\subsection{Neural Network Layers and Methods}
\vspace{-1mm}
PyGSD is built on the existing high-level neural network layer classes from the PyTorch and PyG ecosystems. We define neural network layers for signed and directed networks based on more than 10 architectures.  The constructors of these layers use type hinting to enable the setup of the hyperparameters. Users can
build upon these layers and construct their own model, depending on the task at hand. Besides, our various auxiliary layers could be helpful in building new layers.

\looseness=-1 Building upon the layers, we also implement the full methods from various research papers, so that
end-users can directly utilize the full models. 
For example, users can either employ the \textit{MagNetConv} layer and construct their own full architecture, or they can directly call \textit{MagNet\_node\_classification} or \textit{MagNet\_link\_prediction} if they are interested in applying MagNet \cite{zhang2021magnet} to the node classification or the link prediction task, respectively. A framework diagram is provided in Fig.~\ref{fig:pygsd_diagram} with a logo in Fig.~\ref{fig:pygsd_logo}.
\vspace{-2mm}
\subsection{Data Structures}
\subsubsection{Network Generators and Data Classes}
\vspace{-1mm}
\looseness=-1 PyGSD includes synthetic network generators for models described in Sec.~\ref{subsec:synthetic_data}, i.e., SSBMs and Pol-SSBMs from \cite{he2022sssnet}, DSBMs from \cite{he2022digrac}, as well as SDSBMs from \cite{he2022msgnn}.
To accommodate the distinct features of signed and directed networks, based on the \textit{Data} class from PyG \cite{pytorch_geometric}, our library introduces two data classes for signed and directed networks, respectively. These classes (called \textit{SignedData} and \textit{DirectedData}, respectively) provide properties for checking whether a network is signed (or directed, respectively), and can be initialized to obtain custom data objects with either \textit{edge\_index} only, \textit{edge\_index} together with \textit{edge\_weight}, or the sparse adjacency matrix. Such a data object can also inherit attributes from another data object, where edge attributes are stored in COO format.  

We also provide class functions to construct node features based on edge information. For instance, we can set the node feature matrix as the stacked leading eigenvectors of the regularized adjacency matrix for a signed network as in \cite{he2022sssnet}, or as the stacked real and imaginary parts of top eigenvectors for a Hermitian matrix \cite{cucuringu2020hermitian} constructed from the directed network adjacency matrix as in \cite{he2022digrac}.  

\vspace{-1mm}
\subsubsection{Data Loaders and Splitters}
Our library provides easy-to-use data loaders for real-world data sets for signed and directed networks. The loaded data set is then transformed into a custom data object designed in this library. 
To facilitate downstream tasks, we also provide splitters on the data objects, with only the training set required to be nonempty. Our node splitter generates masks for training, validation, test and seed nodes, where seed nodes are a portion of the training set; such masks can be useful in e.g., the semi-supervised setting of \cite{he2022sssnet}. 
Our link split utility function divides edges into training, validation and test groups, depending on the specific downstream task. The link splitter could be applied to any task mentioned in Sec.~\ref{subsec:link_pred}, and includes an option to keep all edges in the minimum spanning tree in the training set to maintain connectivity for message passing. It also produces more general tools for tasks related to both signs and directionality. 
The splitters can either be called separately or as a data class function.
\vspace{-2mm}
\subsection{Task-Specific Evaluations and Utilities}
\vspace{-1mm}
\looseness=-1 Tasks on signed and directed networks may have custom loss functions and evaluation metrics, such as the probabilistic normalized cut loss and the unhappy ratio introduced in \cite{he2022sssnet}, as well as the probabilistic imbalance objectives introduced in \cite{he2022digrac}. Therefore, PyGSD implements these custom loss functions and evaluation metrics to enable researchers to utilize these functionalities easily. 
Our library also provides some utility functions tailored to our implemented research papers, so that end-users are able to reproduce the whole process of a paper including data preprocessing, training, and testing.
\vspace{-2mm}
\subsection{Maintaining the Library} 
\vspace{-1mm}
\looseness=-1 We maintain our library via open-source code, public releases, automatically updated documentation, code examples, continuous integration, and unit tests with almost \emph{100\%} test coverage. Since its first release in February 2022, the \textit{GitHub} repository has to date attracted 100+ stars.

\textbf{Open-Source library and Public Releases.} The source code of our library is publicly available on \textit{GitHub} under the MIT license. 
The code repository provides contributing guidelines, issue templates and test instructions, which enables the public to contribute to the library and report any problems. The public releases of the library are also made available on the \textit{Python Package Index (PyPI)}. Installation is thus possible via the \textit{pip} command using the terminal.

\looseness=-1 \textbf{Documentation.} We release our software with publicly available documentation at \url{https://pytorch-geometric-signed-directed.readthedocs.io/}, which is automatically built and updated as we modify the main branch of the \textit{GitHub} repository. Our documentation covers signed and directed GNN layers and the full methods in various research papers, data classes specifically designed for signed and directed networks, synthetic data generators, data loaders for real-world data sets, node-level and edge-level splitters, as well as task-specific evaluation metrics, loss and utility functions. The documentation also includes an in-depth installation guide and a tour of external resources.

\textbf{Code Examples.}
In our \textit{GitHub} repository, we present code examples for all papers whose methods have been implemented in our library. Our documentation also introduces examples on data structures, data loaders, and splitters. Some case study examples are provided in Appendix~\ref{appendix:case_study}.

\textbf{Continuous Integration.} We use the freely available \textit{GitHub Actions} to provide continuous integration for our library. When the code is updated on the main branch of the \textit{GitHub} repository or there is a pull request to the main branch, the build process is triggered and the library is deployed on \textit{Linux}, \textit{Windows}, and \textit{macOS} virtual machines, for Python versions 3.7, 3.8 and 3.9. 

\textbf{Unit Tests and Code Coverage.} We provide unit tests for all signed and directed graph neural network layers, task-specific loss functions, evaluation metrics, utility functions, data classes and data loaders. These unit tests can either be executed locally using the source code, or executed through \textit{GitHub  Actions} when the continuous integration process is triggered. When unit tests are running a coverage report by \textit{CodeCov} is generated. We maintain a high test coverage rate of almost  100\%.
\vspace{-4mm}
\section{Comparison with Existing Software} \label{sec:comp} 
\vspace{-1mm}
Similar to existing deep learning software on graphs, which are usually based on machine learning frameworks such as TensorFlow (TF) \cite{abadi2016tensorflow}, PyTorch (PT) \cite{pytorch}, MxNet (MX) \cite{mxnet} and JAX \cite{jax}, PyGSD is built on the PyTorch ecosystem. We summarize the characteristics of related deep learning libraries on graphs in Table \ref{tab:deep_lib_comparison}, comparing frameworks based on the backend, the purpose, whether to contain synthetic data generators, whether to contain dedicated data splitters like our edge or node splitters, and whether to contain dedicated objective functions and evaluation metrics. All libraries listed except OpenNE \cite{openne} provide GPU support. Compared to existing libraries, PyGSD contains various signed GNNs, while PyG \cite{pytorch_geometric} contains only one signed GNN implementation for SGCN~\cite{Derr}, and the other comparison libraries do not contain signed GNNs. To the best of our knowledge, the proposed software is the \emph{first} deep learning library which consists of GNN methods specifically designed for general signed and directed networks with GPU acceleration. Our directed network GNNs are not simple extension to existing GNNs that could be applied to directed networks. In contrast, the directed network method \cite{thost2021directed} contained in CogDL \cite{cen2021cogdl} requires the directed network to be acyclic, which is not generally satisfied.
StellarGraph \cite{StellarGraph} has a GraphSAGE~\cite{hamilton2017inductive} version for directed networks, which is an extension to the original GraphSAGE where in-node and out-node neighborhoods are separately sampled and have different weights. 

While based on PyG, PyGSD contains many new ideas.
In addition to signed and directed GNNs 
PyGSD introduces novel data classes, data loaders, as well as data splitters which are specific to tasks in signed and directed networks. 
PyGSD also includes objective functions and evaluation metrics that are designed for signed and directed graphs, which are not included in PyG. Also our novel synthetic data generators for signed and directed networks are not available in PyG.

\begin{table}[ht]
\vspace{-5.5mm}
\setlength\tabcolsep{1.5pt}
\centering
\caption{Comparison of related open-source libraries for deep learning on graphs. }\label{tab:deep_lib_comparison}
\resizebox{\linewidth}{!}{\begin{tabular}{lcccccccc}
\toprule
Library & Backend          & Purpose&Synthetic Data Generators&Dedicated Data Splitters&Dedicated Objectives/Evaluation Metrics  \\
\midrule
OpenNE  \cite{openne} &Custom      &  General Network Embedding &\xmark&\xmark&\xmark    \\
CogDL   \cite{cen2021cogdl} &PT       &  General GNN Tools &\xmark&\xmark&\xmark    \\
Spektral   \cite{spektral} &TF    &  General GNN Tools   &\xmark&\xmark&\xmark  \\
TF Geometric  \cite{hu2021efficient} &TF&        General GNN Tools \& Methods    &\xmark&\xmark&\xmark\\
StellarGraph   \cite{StellarGraph} &TF       &   General Graph Algorithms &\xmark&\cmark&\xmark \\
DGL   \cite{dlg} &TF/PT/MX       & General GNN Tools \& Methods&\xmark&\xmark&\xmark      \\
DIG   \cite{liu2021dig} &PT       & General GNN Tools \& Methods &\xmark&\xmark&\xmark   \\
Jraph  \cite{jraph2020github}& JAX       &  General GNN Tools \& Methods  &\xmark&\xmark&\xmark    \\
Graph-Learn \cite{zhu2019aligraph}  &TF/PT        &        Large-Scale GNN Tools \& Methods&\xmark&\xmark&\cmark  \\
PyG Temporal \cite{rozemberczki2021pytorch}&PT&           Temporal GNN Tools \& Methods &\xmark&\cmark&\xmark       \\
PyG \cite{pytorch_geometric} &PT&       General GNN Tools \& Methods&\xmark&\xmark&\xmark      \\
\hline
\textbf{Our Work} &PT&   Signed/ Directed GNN Tools \& Methods &\cmark&\cmark&\cmark\\
\bottomrule
\end{tabular}}
\vspace{-3mm}
\end{table}

There are related open-source Python libraries for signed/directed networks that do not contain GNN methods. Python-igraph \cite{csardi2006igraph} contains network analysis tools;
NetworkX \cite{hagberg2008exploring} is a package for complex networks; 
NetworkKit \cite{staudt2016networkit} is a open-source toolkit for large-scale network analysis; 
SigNet\footnote{\url{https://github.com/alan-turing-institute/signet}} \cite{SPONGE_AISTATS_2019} contains spectral analysis methods on signed networks; 
CDLIB \cite{rossetti2019cdlib} allows to extract, compare and evaluate communities from complex networks;
EvalNE \cite{mara2022evalne} is designed for assessing and comparing the performance of network embedding methods on various downstream tasks.

\vspace{-4mm}
\section{Experimental Evaluation}
\vspace{-1mm}
\label{experiments}
To demonstrate that the implemented methods can reproduce the performance in the original papers we evaluate the task performance of all GNNs implemented in the library. We report the mean and one standard deviation for each result, where \textit{$\pm0.0$} indicates that the standard deviation is less than 0.05\%. Implementation details and average runtime are provided in Appendix~\ref{appendix:implementation}. Our implementations generally reproduce the results in the original papers, but slight differences (negligible considering standard deviations) may occur due to data splits or random seeds. The runtime reports also validate our implementation efficiency, showing that our implementations consume comparable or shorter runtime compared to the original implementations. In addition, we provide insights into which methods are preferred for various tasks.
\begin{figure}[ht]
\centering   
\vspace{-2mm}
\begin{subfigure}[ht]{0.52\linewidth}
      \centering  
      \includegraphics[width=0.8\linewidth,trim=1cm 0cm 0.5cm 1.7cm,clip]{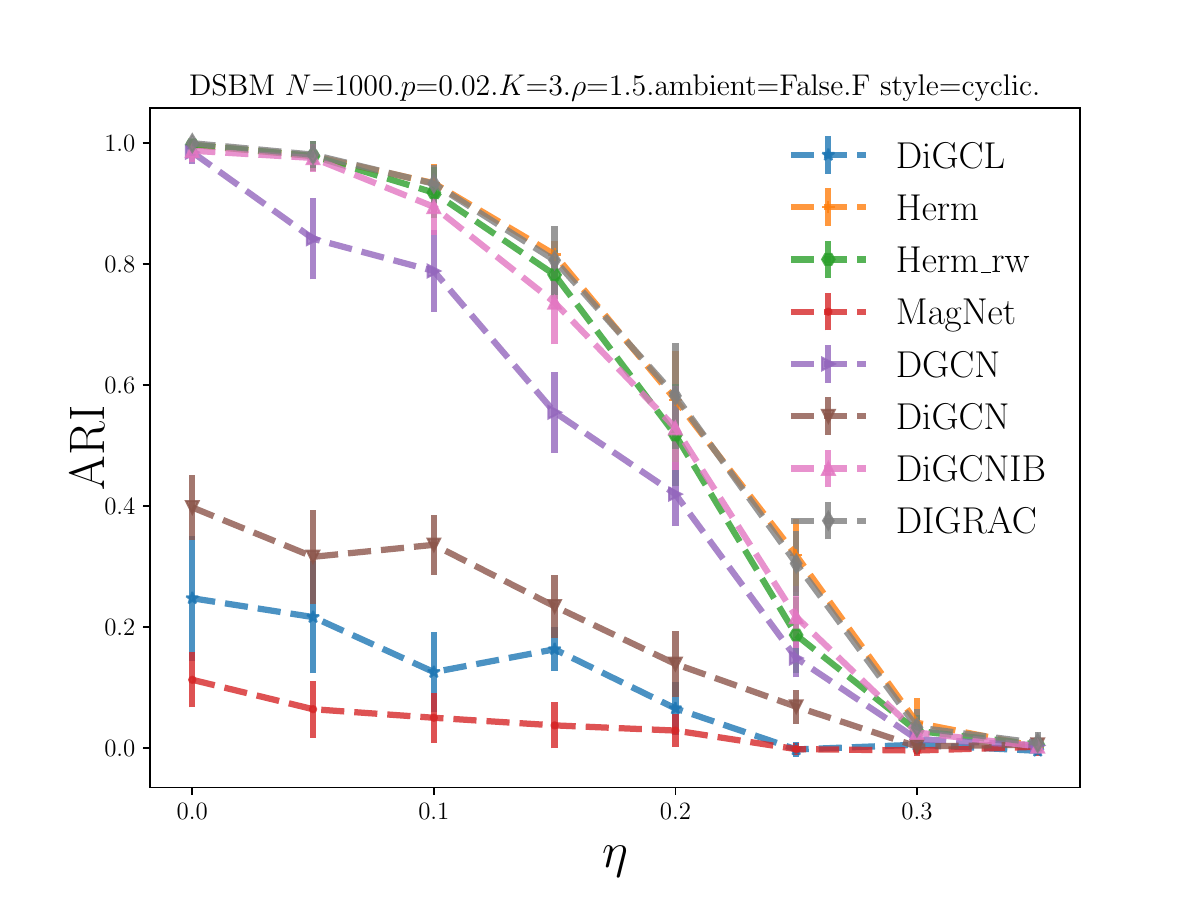}
      \vspace{-2mm}
      \caption{A DSBM model with 3 clusters and edge density 0.02.} 
      \label{fig:DIGRAC}
\end{subfigure}
\begin{subfigure}[ht]{0.47\linewidth}
      \centering
      \includegraphics[width=0.9\linewidth,trim=0cm 0cm 0.5cm 1.8cm,clip]{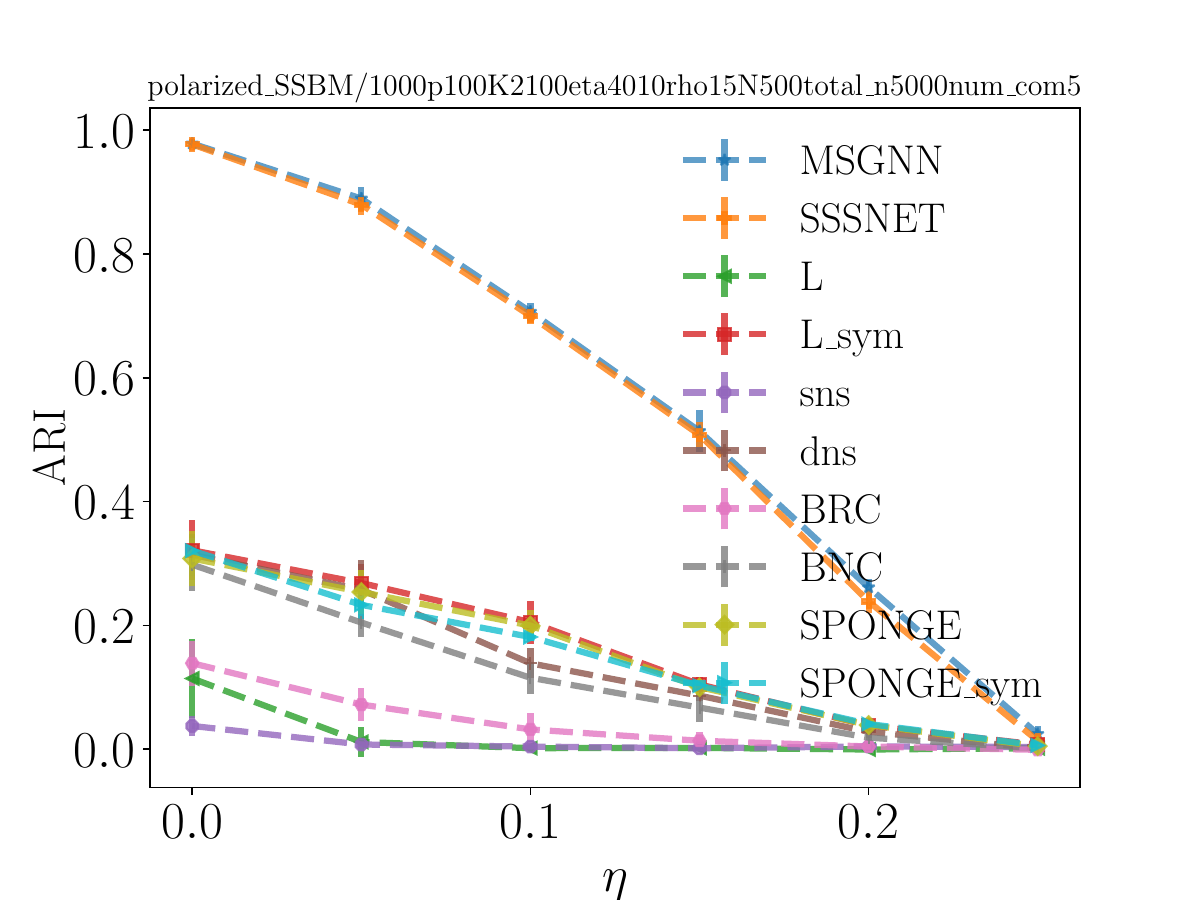}
      \vspace{-2mm}
      \caption{A polarized SSBM model with 5 SSBM blocks.}
      \label{fig:SSSNET}
\end{subfigure}

\begin{subfigure}[ht]{0.489\linewidth}
      \centering  
      \includegraphics[width=0.9\linewidth,trim=0.5cm 0cm 0.5cm 1.71cm,clip]{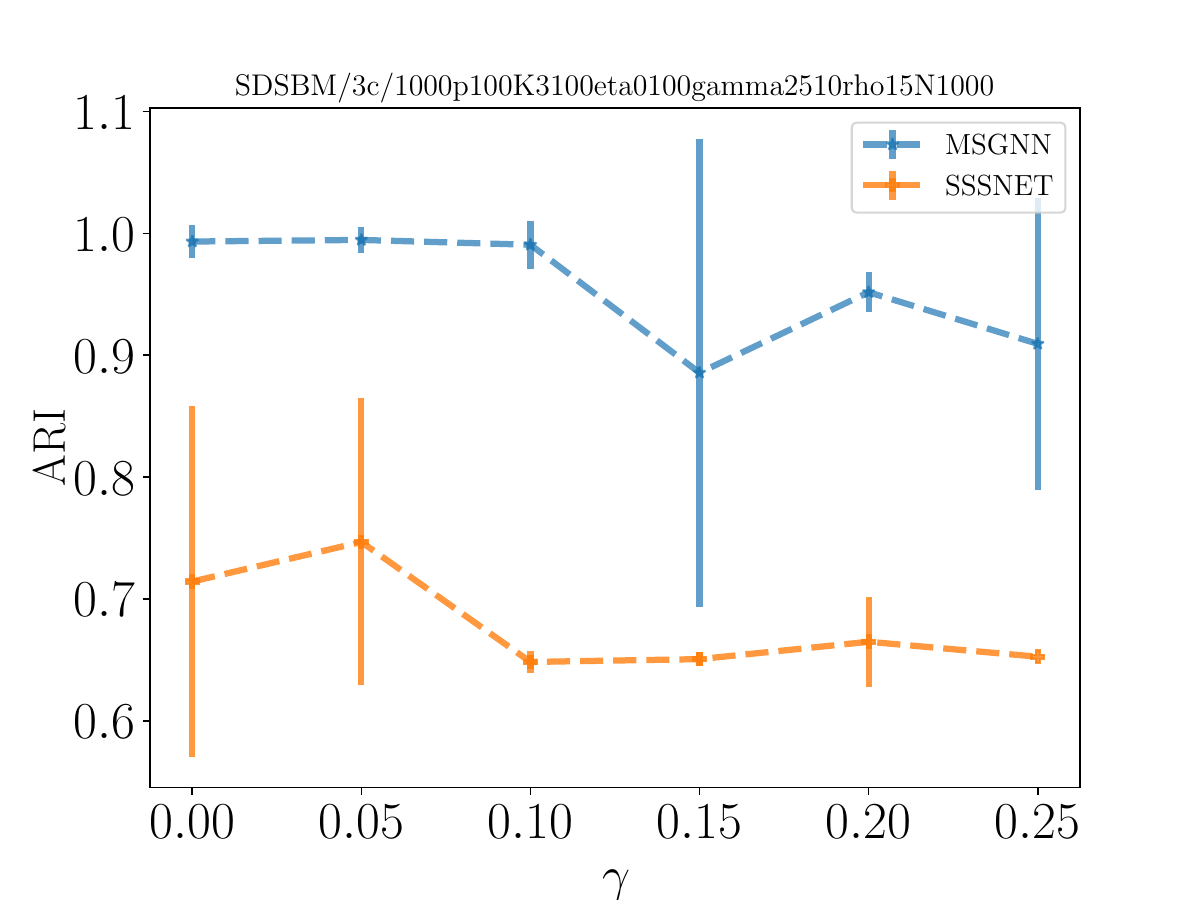}
      \vspace{-2mm}
      \caption{An SDSBM model with meta-graph $\mathbf{F}_1(\gamma)$.} 
      \label{fig:SDSBM_res1}
\end{subfigure}
\begin{subfigure}[ht]{0.489\linewidth}
      \centering  
      \includegraphics[width=0.9\linewidth,trim=0.5cm 0cm 0.5cm 1.71cm,clip]{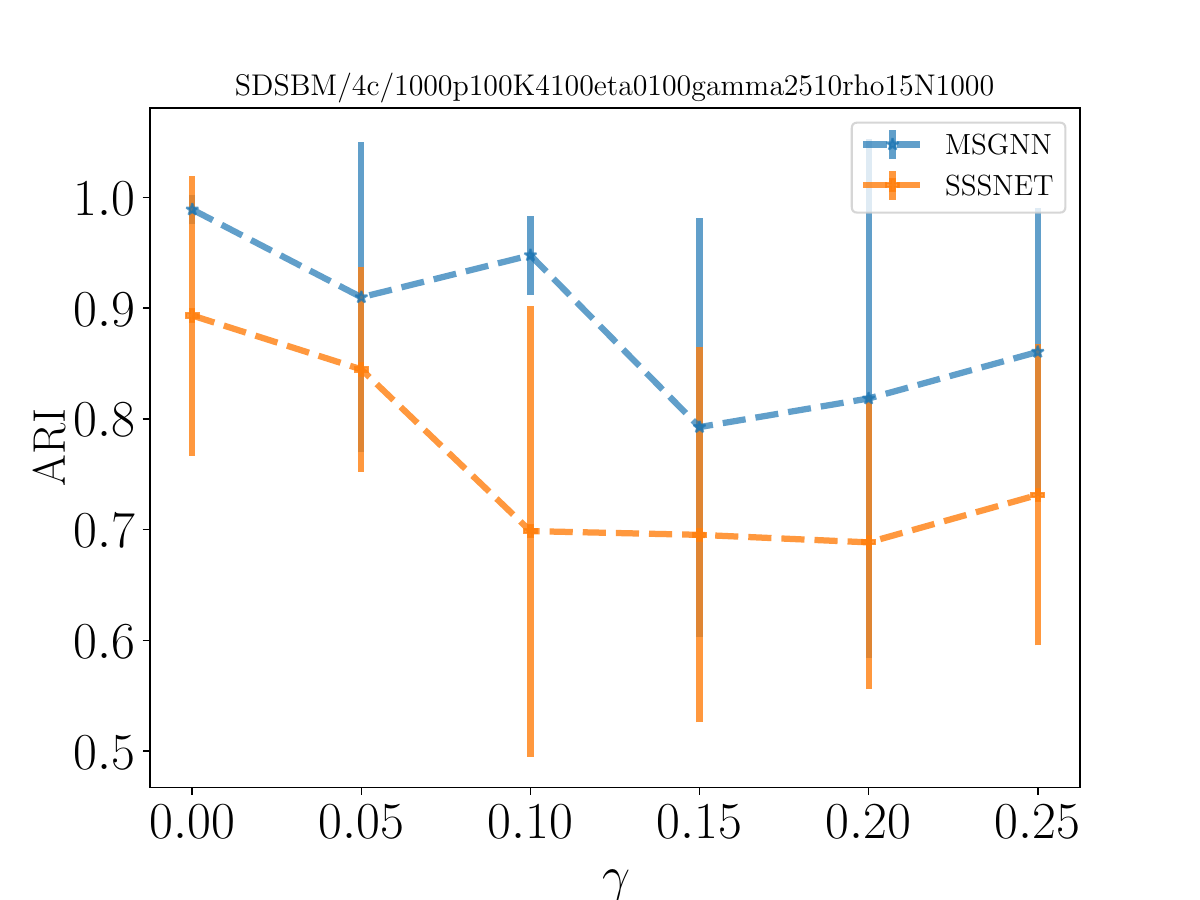}
      \vspace{-2mm}
      \caption{An SDSBM model with meta-graph $\mathbf{F}_2(\gamma)$.} 
      \label{fig:SDSBM_res2}
\end{subfigure}
\vspace{-2mm}
\caption{Test ARI results averaged over 10 runs. Error bars indicate one standard deviation.}
\vspace{-5mm}
\end{figure}

\textbf{Node Clustering.}
Figure~\ref{fig:DIGRAC} reproduces test ARI for the directed unsigned GNNs together with two Hermitian variants mentioned in Sec.~\ref{sec:methods} on a DSBM( ``cyclic", F, $n=1000, K=3, p=0.02, \rho=1.5$) model for three loss variants. 
Figure~\ref{fig:SSSNET} illustrates test ARI for SSSNET \cite{he2022sssnet}, MSGNN~\cite{he2022msgnn} and some spectral methods mentioned in Sec.~\ref{sec:methods} on a polarized SSBM model \textsc{POL-SSBM}($n=5000, r=5, p=0.1, \rho=1.5$), which is similar to the performance in the original paper.
Figure~\ref{fig:SDSBM_res1} and Figure~\ref{fig:SDSBM_res2} reproduce results from \cite{he2022msgnn} on SSSNET and MSGNN for SDSBM ($\mathbf{F}_1(\gamma), n=1000, p=0.1, \rho=1.5, \eta=0.1$) and SDSBM ($\mathbf{F}_2(\gamma), n=1000, p=0.1, \rho=1.5, \eta=0.1$) models, respectively, with varying $\gamma$ where $\mathbf{F}_1$ and $\mathbf{F}_2$ are given in Appendix~\ref{appendix:implementation}. These methods are all efficient in runtime, and are recommended for relative network clustering tasks.

\begin{table}[ht]
\vspace{-5mm}
\caption{Link sign prediction results (AUC, Macro-F1) for signed networks (\%).}
\vspace{-1mm}
\centering
\small 
\setlength\tabcolsep{3pt}
 \resizebox{0.9\linewidth}{!}{
\begin{tabular}{cccccc}
\toprule
Method & Bitcoin-Alpha & Bitcoin-OTC & Wikirfa & Slashdot & Epinions\\ \midrule
SGCN  & 
(87.5$\pm$0.9, 68.9{$\pm$}0.9) &
(88.8$\pm$0.7, 74.1{$\pm$}1.4) &
(84.7$\pm$0.4, 71.7{$\pm$}0.6) &
(79.8$\pm$0.2, 68.2{$\pm$}0.4) &
(87.5$\pm$0.4, 79.6{$\pm$}0.2) \\
SiGAT & 
(86.5$\pm$2.4, 68.6{$\pm$}2.2) &
(88.9$\pm$0.5, 73.8{$\pm$}0.8) &
(87.6$\pm$0.3, 75.3{$\pm$}0.6) &
(87.4$\pm$0.3, 75.5{$\pm$}0.4) &
(92.4$\pm$0.6, 83.6{$\pm$}0.3) \\
SNEA  & 
(\textbf{91.0$\pm$1.1}, 73.3{$\pm$}1.6) &
(\textbf{91.6$\pm$0.4}, 78.6{$\pm$}0.6) &
(87.8$\pm$1.1, 74.8{$\pm$}1.6) &
(\textbf{90.5$\pm$0.2}, \textbf{79.1{$\pm$}0.2}) &
(\textbf{94.8$\pm$0.1}, 86.3{$\pm$}0.2) \\
SDGNN & 
(88.3$\pm$1.4, \textbf{73.7{$\pm$}1.0}) &
(90.8$\pm$0.8, \textbf{79.6{$\pm$}0.4}) &
(\textbf{89.5$\pm$0.1}, \textbf{78.3{$\pm$}0.2}) &
(89.4$\pm$0.3, 78.8{$\pm$}0.4) &
(94.3$\pm$0.2, \textbf{86.6{$\pm$}0.2}) \\
\bottomrule
\end{tabular}
}
\label{tab:link_sign_result}
\vspace{-2mm}
\end{table}
\textbf{Link Sign Prediction on Signed Graphs.}
We use the five real-world data sets to do link sign prediction on SGCN, SiGAT, SNEA, and SDGNN. Table~\ref{tab:link_sign_result} reproduces link sign prediction performance for these four methods on the five real-world data sets. Considering also the runtime comparison given in Table~\ref{tab:link_sign_runtime}, as well as the data set statistics provided in Table~\ref{tab:loaded_data_sets}, we recommend employing SGCN on smaller networks and SiGAT on networks with larger scales, if speed is the main concern. If performance is the main concern, then we would choose SNEA concerning AUC, while we would pick SDGNN regarding Macro-F1.
\begin{wraptable}{r}{0.73\textwidth}
\caption{Evaluation results for directed unsigned graphs (\%).}
\setlength\tabcolsep{3pt}
\vspace{-2mm}
\begin{center}
\resizebox{\linewidth}{!}{\begin{tabular}{ccccccccc}
\toprule
Task&Method&Cornell&Texas&Wisconsin&CoraML&CiteSeer&Telegram \\
\midrule 
\multirow{5}{*}{Node classification}&
DGCN&62.2$\pm$7.8&72.2$\pm$5.7&66.7$\pm$5.4&\textbf{80.1$\pm$2.1}&\textbf{65.5$\pm$1.4}&89.4$\pm$3.2\\
&DiGCN&52.4$\pm$7.3&64.9$\pm$5.1&61.6$\pm$3.3&79.7$\pm$1.1&63.0$\pm$2.7&70.2$\pm$2.9\\
&DiGCNIB&46.2$\pm$6.0&56.2$\pm$4.6&56.3$\pm$5.0&81.2$\pm$1.8&63.8$\pm$2.3&65.4$\pm$5.1\\
&MagNet&\textbf{70.3$\pm$5.1}&\textbf{82.4$\pm$4.9}&\textbf{82.4$\pm$4.4}&70.0$\pm$1.5&60.4$\pm$1.4&\textbf{91.3$\pm$4.1}\\
&DiGCL&33.5$\pm$7.6&55.1$\pm$9.7&44.3$\pm$5.1&74.7$\pm$2.0&55.2$\pm$3.2&77.9$\pm$3.7\\
\midrule
\multirow{5}{*}{Direction prediction}
&DGCN&82.6$\pm$6.1&79.8$\pm$7.3&84.8$\pm$4.7&85.0$\pm$0.8&84.4$\pm$0.9&96.4$\pm$0.6\\
&DiGCN&75.2$\pm$8.9&78.3$\pm$12.5&81.1$\pm$6.5&79.0$\pm$1.8&77.6$\pm$2.6&79.1$\pm$1.9\\
&DiGCNIB&\textbf{86.1$\pm$5.4}&\textbf{80.6$\pm$8.2}&\textbf{85.0$\pm$5.2}&86.0$\pm$0.5&\textbf{85.6$\pm$0.7}&96.5$\pm$0.5\\
&MagNet&79.8$\pm$9.4&78.9$\pm$8.8&84.1$\pm$4.7&\textbf{86.8$\pm$0.7}&84.6$\pm$0.7&\textbf{97.4$\pm$0.5}\\
\midrule
\multirow{5}{*}{Existence link prediction}
&DGCN&61.6$\pm$4.8&59.3$\pm$5.0&66.3$\pm$5.4&76.8$\pm$2.0&65.6$\pm$2.3&86.3$\pm$0.9\\
&DiGCN&59.9$\pm$5.4&60.8$\pm$7.3&60.1$\pm$4.7&74.5$\pm$1.0&68.5$\pm$1.3&73.9$\pm$2.0\\
&DiGCNIB&66.7$\pm$3.5&\textbf{68.9$\pm$3.9}&\textbf{71.0$\pm$4.1}&\textbf{77.8$\pm$0.6}&\textbf{73.0$\pm$1.2}&84.2$\pm$1.1\\
&MagNet&\textbf{66.9$\pm$4.2}&59.9$\pm$7.4&68.4$\pm$4.2&77.0$\pm$0.9&65.9$\pm$2.3&\textbf{86.9$\pm$0.6}\\
\midrule
\multirow{5}{*}{Three classes link prediction}
&DGCN&\textbf{60.6$\pm$6.8}&61.5$\pm$8.6&\textbf{69.2$\pm$4.6}&69.3$\pm$1.5&62.2$\pm$3.9&81.6$\pm$0.7\\
&DiGCN&51.8$\pm$6.7&56.5$\pm$7.2&49.6$\pm$4.2&65.0$\pm$1.4&54.0$\pm$1.4&63.6$\pm$5.4\\
&DiGCNIB&\textbf{60.6$\pm$5.2}&\textbf{67.2$\pm$5.9}&66.8$\pm$5.8&70.3$\pm$0.8&\textbf{68.5$\pm$1.1}&79.9$\pm$0.6\\
&MagNet&56.0$\pm$10.4&63.2$\pm$6.6&67.5$\pm$4.1&\textbf{70.5$\pm$1.1}&58.9$\pm$3.3&\textbf{83.3$\pm$0.6}\\
\bottomrule
\end{tabular}}
\end{center}
\label{tab:eva_results}
\vspace{-6mm}
\end{wraptable}

\textbf{Node Classification and Edge-Level Tasks on Directed Unsigned Graphs.}
Table~\ref{tab:eva_results} present results on node classification and three link prediction tasks for directed unsigned graphs, for five directed graph methods (DGCN, DiGCN, DiGCNIB, MagNet, and DiGCL), where DiGCNIB is the ``inception block" version of DiGCN introduced in \cite{tong2020digraph}. Given also the runtime comparison in Table~\ref{tab:directed_gcn_time}, and the data set statistics detailed in Table~\ref{tab:loaded_data_sets}, we suggest using DiGCN, when speed is the main concern. If task performance is more important, then we would recommend DiGCNIB as the method of choice for data sets where directionality is largely emphasized (such as the data sets here except CoraML and CiteSeer). On CoraML and CiteSeer, the best-performing method varies case by case.

\begin{wraptable}{r}{0.73\textwidth}
\vspace{-5mm}
\caption{Link prediction test accuracy (\%) comparison for directions (and signs). The link prediction tasks are introduced in Sec.~\ref{subsec:link_pred}.}
\vspace{-1mm}
  \label{tab:link_sign_direction_prediction_performance}
  \centering
  \small
  \resizebox{\linewidth}{!}{\begin{tabular}{cc c c c  c  cccccc}
    \toprule
    Data Set& Link Task&GCN&SGCN&SDGNN&SiGAT&SNEA&SSSNET&MSGNN \\
    \midrule
\multirow{4}{*}{BitCoin-Alpha}&SP&58.0$\pm$2.0&64.7$\pm$0.9&64.5$\pm$1.1&62.9$\pm$0.9&64.1$\pm$1.3&67.4$\pm$1.1&\textbf{71.3$\pm$1.2}\\
&DP&59.6$\pm$1.3&60.4$\pm$1.7&61.5$\pm$1.0&61.9$\pm$1.9&60.9$\pm$1.7&68.1$\pm$2.3&\textbf{72.5$\pm$1.5}\\
&3C&82.0$\pm$0.4&81.4$\pm$0.5&79.2$\pm$0.9&77.1$\pm$0.7&83.2$\pm$0.5&78.3$\pm$4.7&\textbf{84.4$\pm$0.6}\\
&4C&42.4$\pm$2.2&51.1$\pm$0.8&52.5$\pm$1.1&49.3$\pm$0.7&52.4$\pm$1.8&54.3$\pm$2.9&\textbf{58.5$\pm$0.7}\\
&5C&80.0$\pm$0.4&79.5$\pm$0.3&78.2$\pm$0.5&76.5$\pm$0.3&81.1$\pm$0.3&77.9$\pm$0.3&\textbf{81.9$\pm$0.9}\\
\midrule
\multirow{4}{*}{BitCoin-OTC}&SP&55.3$\pm$1.2&65.6$\pm$0.9&65.3$\pm$1.2&62.8$\pm$1.3&67.7$\pm$0.5&70.1$\pm$1.2&\textbf{73.0$\pm$1.4}\\
&DP&55.3$\pm$2.3&63.8$\pm$1.2&63.2$\pm$1.5&64.0$\pm$2.0&65.3$\pm$1.2&69.6$\pm$1.0&\textbf{71.8$\pm$1.1}\\
&3C&78.8$\pm$0.8&79.0$\pm$0.7&77.3$\pm$0.7&73.6$\pm$0.7&82.2$\pm$0.4&76.9$\pm$1.1&\textbf{83.3$\pm$0.7}\\
&4C&42.2$\pm$1.2&51.5$\pm$0.4&55.3$\pm$0.8&51.2$\pm$1.8&56.9$\pm$0.7&57.0$\pm$2.0&\textbf{59.8$\pm$0.7}\\
&5C&75.8$\pm$0.5&77.4$\pm$0.7&77.3$\pm$0.8&74.1$\pm$0.5&80.5$\pm$0.5&74.0$\pm$1.6&\textbf{80.9$\pm$0.9}\\
\midrule
\multirow{4}{*}{Slashdot}&SP&78.3$\pm$1.1&74.7$\pm$0.5&74.1$\pm$0.7&64.0$\pm$1.3&70.6$\pm$1.0&86.6$\pm$2.2&\textbf{92.4$\pm$0.2}\\
&DP&79.1$\pm$0.2&74.8$\pm$0.9&74.2$\pm$1.4&62.8$\pm$0.9&71.1$\pm$1.1&87.8$\pm$1.0&\textbf{93.1$\pm$0.1}\\
&3C&74.7$\pm$1.0&69.7$\pm$0.3&66.3$\pm$1.8&49.1$\pm$1.2&72.5$\pm$0.7&79.3$\pm$1.2&\textbf{86.1$\pm$0.3}\\
&4C&60.0$\pm$1.1&63.2$\pm$0.3&64.0$\pm$0.7&53.4$\pm$0.2&60.5$\pm$0.6&72.7$\pm$0.6&\textbf{78.2$\pm$0.3}\\
&5C&65.3$\pm$0.2&64.4$\pm$0.3&62.6$\pm$2.0&44.4$\pm$1.4&66.4$\pm$0.5&70.4$\pm$0.7&\textbf{76.8$\pm$0.6}\\
\midrule
\multirow{4}{*}{Epinions}&SP&68.6$\pm$1.2&62.9$\pm$0.5&67.7$\pm$0.8&63.6$\pm$0.5&66.5$\pm$1.0&78.5$\pm$2.1&\textbf{85.4$\pm$0.5}\\
&DP&67.6$\pm$1.0&61.7$\pm$0.5&67.9$\pm$0.6&63.6$\pm$0.8&66.4$\pm$1.2&73.9$\pm$6.2&\textbf{86.3$\pm$0.3}\\
&3C&73.6$\pm$1.1&70.3$\pm$0.8&73.2$\pm$0.8&52.3$\pm$1.3&72.8$\pm$0.2&72.7$\pm$2.0&\textbf{83.1$\pm$0.5}\\
&4C&61.3$\pm$1.2&66.7$\pm$1.2&71.0$\pm$0.6&62.3$\pm$0.5&69.5$\pm$0.7&70.2$\pm$5.2&\textbf{78.7$\pm$0.9}\\
&5C&70.0$\pm$0.7&73.5$\pm$0.8&76.6$\pm$0.7&52.9$\pm$0.7&74.2$\pm$0.1&70.3$\pm$4.6&\textbf{80.5$\pm$0.5}\\
\midrule
\multirow{4}{*}{FiLL (avg.)}&SP&87.7$\pm$0.0&88.4$\pm$0.0&82.0$\pm$0.3&76.9$\pm$0.1&90.0$\pm$0.0&88.7$\pm$0.3&\textbf{90.8$\pm$0.0}\\
&DP&87.7$\pm$0.0&88.5$\pm$0.1&82.0$\pm$0.2&76.9$\pm$0.1&90.0$\pm$0.0&88.8$\pm$0.3&\textbf{90.9$\pm$0.0}\\
&3C&61.7$\pm$0.1&63.0$\pm$0.1&59.3$\pm$0.0&55.3$\pm$0.1&64.3$\pm$0.1&62.2$\pm$0.3&\textbf{66.1$\pm$0.1}\\
&4C&71.6$\pm$0.1&81.7$\pm$0.0&78.8$\pm$0.1&70.5$\pm$0.1&83.2$\pm$0.1&80.0$\pm$0.3&\textbf{83.3$\pm$0.0}\\
&5C&54.6$\pm$0.1&63.8$\pm$0.0&61.1$\pm$0.1&55.5$\pm$0.1&\textbf{64.8$\pm$0.1}&60.4$\pm$0.4&\textbf{64.8$\pm$0.1}\\
\bottomrule
\end{tabular}}
\vspace{-4mm}
\end{wraptable}
\textbf{Link Prediction on Signed Directed Graphs.}
Table~\ref{tab:link_sign_direction_prediction_performance} reproduces the results on five link prediction tasks introduced in Sec.~\ref{subsec:link_pred}, on the six signed methods implemented in the library (SGCN, SDGNN, SiGAT, SNEA, SSSNET, and MSGNN) and an unsigned baseline GCN, on five real-world signed directed data sets. Taking in addition the runtime comparison in Table~\ref{tab:runtime_link_pred} into account, as well as the data set statistics given by Table~\ref{tab:loaded_data_sets}, we recommend MSGNN as the method of choice. Note that SSSNET is also quite competitive in terms of performance while being fast for large networks. The GCN employed here replaces MSConv in MSGNN with GCNConv from PyG, and additionally inputs absolute values of edge weights for message passing to avoid getting NAN values. We conclude that signed information is indeed helpful when comparing GCN and MSGNN. GCN is generally faster but not by a large margin, while MSGNN is considerably more accurate.
\vspace{-3mm}
\section{Conclusion and Outlook} 
\vspace{-1mm}
\label{sec:conclusion}
We present \textit{PyTorch Geometric Signed Directed} (PyGSD), an easy-to-use and easy-to-adapt end-to-end software package which uses GNNs to address three main tasks in the analysis of signed and directed networks: link prediction, node classification, and node clustering. Our evaluation and discussion of the implemented methods provide insights for users on which methods to pick for various tasks. 
As an extension library to PyG, our proposed PyGSD is open source and comes with an invitation to contribute new methods, data sets, utilities, and case studies. 
In the future, we envisage contributions in the area of scalability and metrics for assessing specific tasks. Our library is related to PyTorch Geometric Temporal, another extension library to PyG developed for the analysis of temporal networks. In future works, these two software packages are likely to grow closer together. 
\clearpage
\section*{Author Contributions}
Y.H.: Conceptualization, Data curation, Formal analysis, Funding acquisition, Investigation, Methodology, Project administration, Software, Visualization, Writing - original draft; X.Z.: Data curation, Software, Writing - review \& editing; J.H: Data curation, Software, Writing - review \& editing; B.R.: Writing - review \& editing; M.C.: Conceptualization, Data curation, Formal analysis, Funding acquisition, Methodology, Project administration, Resources, Supervision, Visualization, Writing - review \& editing; G.R.: Conceptualization, Formal analysis, Funding acquisition, Methodology, Project administration, Supervision, Validation, Writing - review \& editing.

\section*{Acknowledgements} 
Y.H. is supported by a Clarendon scholarship. 
G.R. is  supported in part by EPSRC grants EP/T018445/1,  EP/W037211/1 and  EP/R018472/1. 
M.C. acknowledges support from the EPSRC grants EP/N510129/1
and EP/W037211/1 at The Alan Turing Institute.
\bibliographystyle{plain}
\bibliography{log_2023}
\clearpage
\appendix
\section{Logo and Diagram}
\begin{figure}[h]
    \centering
    \includegraphics[width=0.7\linewidth]{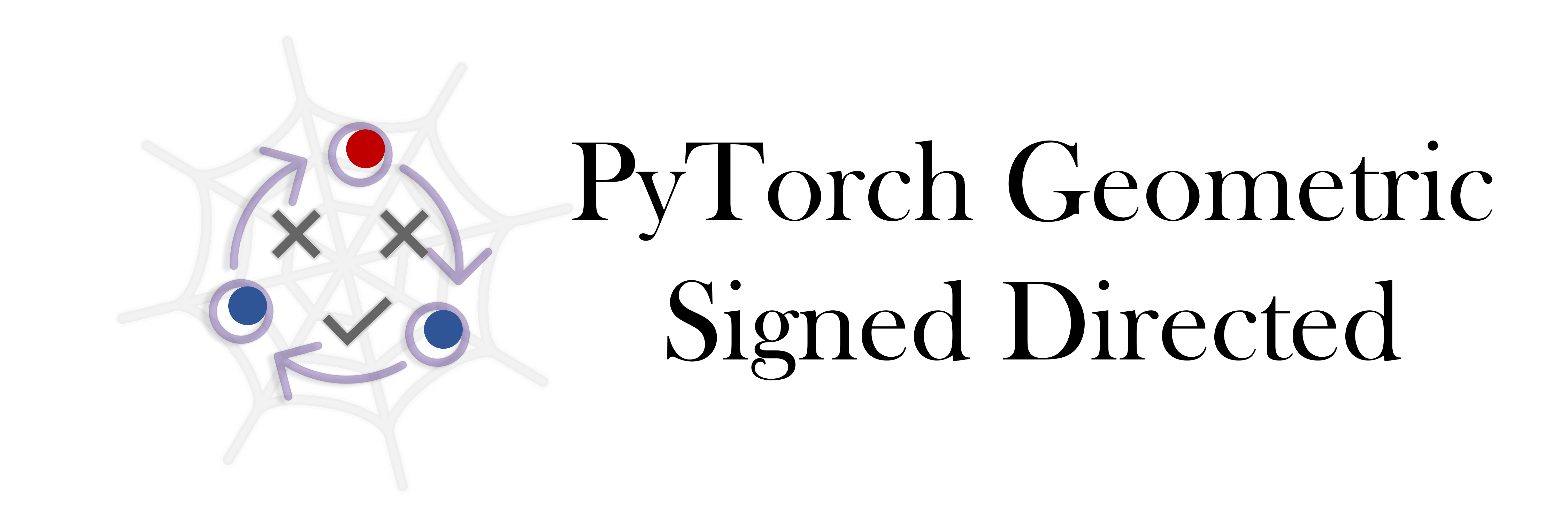}
    \caption{PyGSD Logo.}
    \label{fig:pygsd_logo}
\end{figure}
\begin{figure}[h]
    \centering
    \includegraphics[width=\linewidth]{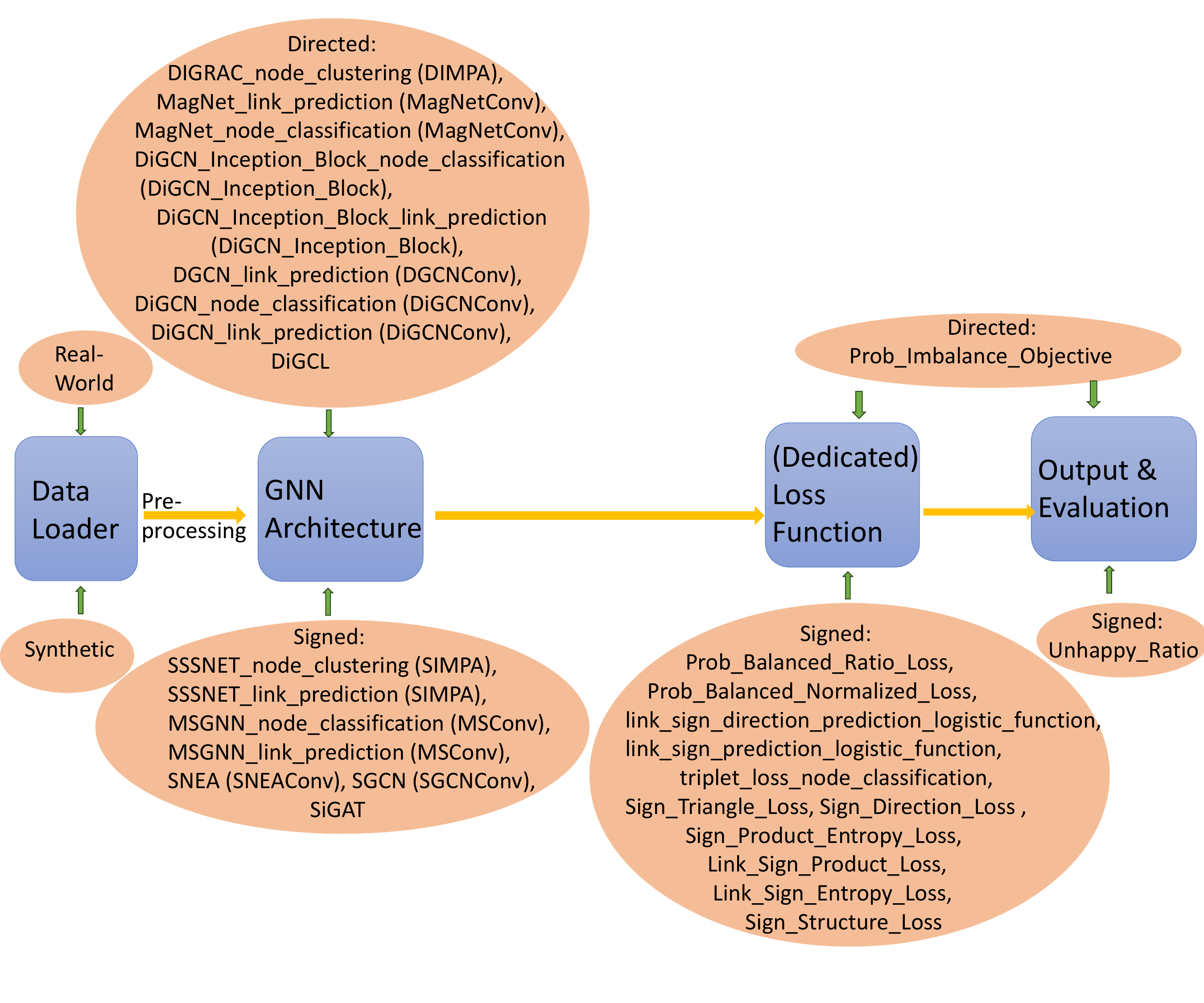}
    \caption{PyGSD framework overview: starting from our data loaders, input data is loaded into the GNN architecture (where the layers inside brackets could be treated individually to construct new architectures), then (dedicated) loss functions are employed, and finally we test the performance.}
    \label{fig:pygsd_diagram}
\end{figure}
\section{Case Study Examples}
\label{appendix:case_study}

Here we first  provide a case study for clustering in signed networks, and then a case study for link prediction in directed networks.

\subsection{Case Study on Signed Networks}
In this subsection, we overview a simple end-to-end machine learning pipeline  for signed networks designed with \textit{PyTorch Geometric Signed Directed} (PyGSD). These three code snippets -- Listings \ref{code:split_example}, \ref{code:training} and \ref{code:evaluation} -- solve the signed clustering problem on a Signed Stochastic Block Model of clustering the nodes in the signed network into five groups.  The pipeline consists of data preparation (Listing \ref{code:split_example}), model definition, training, and evaluation phases (Listing \ref{code:training}). Listing \ref{code:evaluation} illustrates how to run experiments and output results. We explain the three listings in turn.

\medskip 
\begin{code}
\begin{minted}[linenos,
xleftmargin=0.2cm,numbersep=3pt,frame=lines]{python}
from sklearn.metrics import adjusted_rand_score
import scipy.sparse as sp
import torch
from torch_geometric_signed_directed.nn import \
    SSSNET_node_clustering as SSSNET
from torch_geometric_signed_directed.data import \
    SignedData, SSBM
from torch_geometric_signed_directed.utils import \
    (Prob_Balanced_Normalized_Loss, 
extract_network, triplet_loss_node_classification)

device = torch.device('cuda' if \
torch.cuda.is_available() else 'cpu')

num_classes = 5
num_nodes = 1000
(A_p_scipy, A_n_scipy), labels = SSBM(num_nodes, \ 
num_classes, 0.1, 0.1)
A = A_p_scipy - A_n_scipy
A, labels = extract_network(A=A, labels=labels)
data = SignedData(A=A, y=torch.LongTensor(labels))
data.set_spectral_adjacency_reg_features(num_classes)
data.node_split(train_size_per_class=0.8, \ 
val_size_per_class=0.1, \ 
test_size_per_class=0.1, seed_size_per_class=0.1)
data.separate_positive_negative()
data = data.to(device)
loss_func_ce = torch.nn.NLLLoss()
model = SSSNET(nfeat=data.x.shape[1], dropout=0.5,  
hop=2, fill_value=0.5, hidden=32, nclass=num_classes).to(device)
\end{minted}
\captionof{listing}{Preparation, data loading, node splitting, as well as loss and model initialization.}\label{code:split_example}
\end{code}
\subsubsection{Preparation, Data Loading and Splitting, Loss and Model Initialization} 
In Listing \ref{code:split_example}, as a first step, we import the essentials (lines 1-10). We then define the device to be used (lines 12-13). 
After that, we define default values to be used in the network generation process, generate a synthetic network and extract its 
largest connected component (lines 15-20); we use the SignedData class (line 21). As no node features are available initially, we use the \textit{data.set\_spectral\_adjacency\_reg\_features()} class method to set up the node feature matrix (line 22).

We then create a train-validation-test-seed split of the node set by using the node splitting function and calculate separate positive and negative parts of the signed network to be stored inside the data object (lines 23-26). 
We then move the data object to the device (line 27). Finally, we initialize the cross-entropy loss function (line 28), construct the GNN model and map it to the device (lines 29-30). 

\medskip
\begin{code}
\begin{minted}[linenos,
xleftmargin=0.2cm,numbersep=3pt,frame=lines]{python}
def train(features, edge_index_p, edge_weight_p,
                edge_index_n, edge_weight_n, mask, seed_mask,
                loss_func_pbnc, y):
    model.train()
    Z, log_prob, _, prob = model(edge_index_p, edge_weight_p,
                edge_index_n, edge_weight_n, features)
    loss_pbnc = loss_func_pbnc(prob[mask])
    loss_triplet = triplet_loss_node_classification(
    y=y[seed_mask], Z=Z[seed_mask], n_sample=500, thre=0.1)
    loss_ce = loss_func_ce(log_prob[seed_mask], y[seed_mask])
    loss = 50*(loss_ce + 0.1*loss_triplet) + loss_pbnc
    optimizer.zero_grad()
    loss.backward()
    optimizer.step()
    train_ari = adjusted_rand_score(y[mask].cpu(),
    (torch.argmax(prob, dim=1)).cpu()[mask])
    return loss.detach().item(), train_ari

def test(features, edge_index_p, edge_weight_p,
                edge_index_n, edge_weight_n, mask, y):
    model.eval()
    with torch.no_grad():
        _, _, _, prob = model(edge_index_p, edge_weight_p,
                edge_index_n, edge_weight_n, features)
    test_ari = adjusted_rand_score(y[mask].cpu(),
    (torch.argmax(prob, dim=1)).cpu()[mask])
    return test_ari
\end{minted}
\captionof{listing}{Defining the training and evaluation functions.}\label{code:training}
\end{code}
\subsubsection{Defining Functions for Training and Evaluation} In Listing  \ref{code:training}, we define the training and evaluation functions. Setting the model to be trainable (line 4), we obtain the node embedding matrix \textit{Z} and cluster assignment probabilities \textit{prob} and its logarithm \textit{log\_prob} with a forward pass of the model instance (lines 5-6). We then obtain the probabilistic balanced normalized cut loss (line 7), triplet loss (lines 8-9), and cross entropy loss (line 10) values. The weighted sum of the three losses serves as the training loss value (line 11). 

We then backpropagate and update the model parameters (lines 12-14). After that, we calculate the ARI of the training nodes (lines 15-16). Finally, we return the loss value and training ARI (line 17).

For evaluation (function \textit{test()}), we do not set the model to be trainable (lines 21-22). With a forward pass (lines 23-24), we obtain the probability assignment matrix. Taking argmax for the probabilities, we obtain test ARI (lines 25-26) and return the result (line 27).

\begin{code}
\begin{minted}[linenos,
xleftmargin=0.2cm,numbersep=3pt,frame=lines]{python}
for split in range(data.train_mask.shape[1]):
    optimizer = torch.optim.Adam(model.parameters(),
    lr=0.01, weight_decay=0.0005)
    train_index = data.train_mask[:, split].cpu().numpy()
    val_index = data.val_mask[:, split]
    test_index = data.test_mask[:, split]
    seed_index = data.seed_mask[:, split]
    loss_func_pbnc = Prob_Balanced_Normalized_Loss(
    A_p=sp.csr_matrix(data.A_p)[train_index][:, train_index], 
    A_n=sp.csr_matrix(data.A_n)[train_index][:, train_index])
    for epoch in range(300):
        train_loss, train_ari = train(data.x,
        data.edge_index_p,
        data.edge_weight_p, data.edge_index_n,
        data.edge_weight_n, train_index,
        seed_index, loss_func_pbnc, data.y)
        Val_ari = test(data.x, data.edge_index_p,
        data.edge_weight_p, data.edge_index_n,
        data.edge_weight_n, val_index, data.y)
        print(f'Split: {split:02d}, Epoch: {epoch:03d}, 
        Train_Loss: {train_loss:.4f},
        Train_ARI: {train_ari:.4f},
        Val_ARI: {Val_ari:.4f}')
    
    test_ari = test(data.x, data.edge_index_p, 
    data.edge_weight_p, data.edge_index_n,
    data.edge_weight_n, test_index, data.y)
    print(f'Split: {split:02d}, Test_ARI: {test_ari:.4f}')
    model._reset_parameters_undirected()
\end{minted}
\captionof{listing}{Running the experiment and printing the results.}  \label{code:evaluation}
\end{code}

\subsubsection{Running the Experiment} 
We run the actual experiments in Listing  \ref{code:evaluation}. For each of the data splits (line 1), we first initialize the Adam optimizer \cite{kingma_adam_2014} (lines 2-3). We then obtain the data split indices (lines 4-7), initialize the self-supervised loss function (lines 8-10), and start the training process (line 11). 

For each epoch,  we apply the training function to obtain training loss and ARI score (lines 12-16), then evaluate with the \textit{test()} function on validation nodes (lines 17-19).  We then print the training and validation results (lines 20-23). 
After training, we obtain the test performance (lines 25-27) and print some logs (line 28). Finally, we reset the model parameters (line 29) and iterate to the next data split loop.

\subsection{Case Study on Directed Networks}
In this subsection, we overview a simple end-to-end machine learning pipeline  for directed networks designed with PyGSD. These code snippets solve a link direction prediction problem on a real-world data set. We skip the actual training loop here; 
full examples can be found in the code repository.
\begin{code}
\begin{minted}[linenos,
xleftmargin=0.2cm,numbersep=3pt,frame=lines]{python}
from sklearn.metrics import accuracy_score
import torch

from torch_geometric_signed_directed.utils import \ 
link_class_split, in_out_degree
from torch_geometric_signed_directed.nn.directed import \ 
MagNet_link_prediction
from torch_geometric_signed_directed.data import \ 
load_directed_real_data

device = torch.device('cuda' if \
torch.cuda.is_available() else 'cpu')

data = load_directed_real_data(dataset='webkb', 
root=path, name='cornell').to(device)
link_data = link_class_split(data, prob_val=0.15, 
prob_test=0.05, task = 'direction', device=device)
model = MagNet_link_prediction(q=0.25, K=1, num_features=2, 
hidden=16, label_dim=2).to(device)
criterion = torch.nn.NLLLoss()
\end{minted}
\captionof{listing}{Preparation, data loading, link splitting, as well as loss and model initialization.}\label{code:split_example2}
\end{code}
\subsubsection{Preparation, Data Loading and Splitting, Loss and Model Initialization} 
In Listing  \ref{code:split_example2}, after importing and defining the device (lines 1-12), we load the \textit{DirectedData} object for the selected data set and map it to the device (lines 14-15). We then create a train-validation-test split of the edge set by using the directed link splitting function (lines 16-17). Finally, we  construct the model instance (lines 18-19), and initialize the cross-entropy loss function (line 20).
\begin{code}
\begin{minted}[linenos,
xleftmargin=0.2cm,numbersep=3pt,frame=lines]{python}
def train(X_real, X_img, y, edge_index,
edge_weight, query_edges):
    model.train()
    out = model(X_real, X_img, edge_index=edge_index, 
                    query_edges=query_edges, 
                    edge_weight=edge_weight)
    loss = criterion(out, y)
    optimizer.zero_grad()
    loss.backward()
    optimizer.step()
    train_acc = accuracy_score(y.cpu(),
    out.max(dim=1)[1].cpu())
    return loss.detach().item(), train_acc

def test(X_real, X_img, y, edge_index, edge_weight, 
query_edges):
    model.eval()
    with torch.no_grad():
        out = model(X_real, X_img, edge_index=edge_index, 
                    query_edges=query_edges, 
                    edge_weight=edge_weight)
    test_acc = accuracy_score(y.cpu(),
    out.max(dim=1)[1].cpu())
    return test_acc
\end{minted}
\captionof{listing}{Defining the training and evaluation functions.}\label{code:training2}
\end{code}
\subsubsection{Defining Functions for Training and Evaluation} In Listing  \ref{code:training2}, we define the training and evaluation functions. Setting the model to be trainable (line 3), we obtain edge class assignment probablities with a forward pass of the model instance (lines 4-6). We then obtain the training loss value (line 7). After that, we backpropagate and update the model parameters (lines 8-10). Then, we calculate the accuracy of the training samples (lines 11-12). Finally, we return the loss value as well as the training accuracy (line 13).

For the evaluation function (function \textit{test()}), we do not set the model to be trainable (lines 17-18). With a forward pass (lines 19-21), we obtain the probability assignment matrix. We then obtain test accuracy (line 22) and return the result (line 23).
\section{Synthetic Data Description}
\label{appendix_sec:synthetic_description}
\subsection{Signed Stochastic Block Model (SSBM)}
\label{subsubsec:ssbm}
A Signed Stochastic Block Model (SSBM) \cite{he2022sssnet} for a network on $n$ nodes  with $K$ blocks (clusters), is constructed as follows: 

1) Assign block sizes $n_0 \le n_1 \le \cdots \le n_{K-1}$ 
with size ratio $\rho \geq 1$, as follows. If $\rho=1$, then the first $K-1$  blocks have the same size $\lfloor n/K \rfloor$,  and the last block has size $ n - (K-1) \lfloor n/K \rfloor$. If $\rho > 1$, we set 
$\rho_0=\rho^{\frac{1}{K-1}}.$
Solving 
$\sum_{i=0}^{K-1}\rho_0^in_0=n$ and taking
integer value gives $n_0=\left\lfloor {n(1-\rho_0)} / {(1-\rho_0^K)}\right\rfloor.$
Further, set $n_i=\lfloor\rho_0n_{i-1}\rfloor,$ for $i=1,\cdots, K-2$ if $K\geq 3,$ and $n_{K-1}=n-\sum_{i=0}^{K-2}n_i.$
Then, the ratio of the size of the largest to the smallest block is roughly $\rho_0^{K-1}=\rho.$ 

2) 
Assign each node to one of $K$ blocks, so that each block has the allocated size. 

3)~For each pair of nodes in the same block, with probability $p_\text{in}$,  create an edge with $+1$ as weight between them, independently of the other potential edges. 

4) For each pair of nodes in different blocks,  with probability $p_\text{out},$  create an edge with $-1$ as weight between them, independently of the other potential edges. 

5) Flip the sign of the across-cluster edges from the previous stage with sign flip probability $\eta_\text{in}$, and $\eta_\text{out}$ for edges within and across clusters, respectively.

\subsection{Polarized SSBMs (POL-SSBM)} 
In a polarized SSBM model \cite{he2022sssnet}, SSBMs are planted in an ambient network; each block of each SSBM is a cluster, and the nodes not assigned to any SSBM form an ambient cluster. 
The
polarized SSBM model
that creates communities of SSBMs, is generated as follows: 

1) Generate an {\ER} 
    graph with $n$ nodes and edge probability $p,$ whose sign is set to $ \pm 1$ with equal probability 0.5. 
    
2) Fix $n_c$ as the number of SSBM communities,  and   calculate community sizes $N_1 \le N_2 \le \cdots \le N_{r},$ for each of the $r$ communities as in Sec.~\ref{subsubsec:ssbm}, such that the ratio of the largest block size to the smallest block size is  
 approximately 
$\rho$, and the total number of nodes in these SSBMs is $N\times n_c.$ 

3)~Generate $r$ SSBM models, each with $K_i=2, i = 1, \ldots,r$ blocks,  
number of nodes according to its community size, with the same edge probability $p$, size ratio $\rho$, and flip probability $\eta$. 

4) Place the SSBM models on disjoint subsets of the whole network; the remaining nodes not part of any SSBM are dubbed
as {\it ambient} nodes. 
The resulting polarized SSBM model is denoted as {\textsc{Pol-SSBM}} ($n,r,p,\rho,\eta,N$).

\subsection{Directed Stochastic Block Model (DSBM)}
A standard directed stochastic blockmodel (DSBM) is often used to represent a  network cluster structure, see for example \cite{malliaros2013clustering}. A DSBM model relies on a meta-graph adjacency matrix $\mathbf{F} = (\mathbf{F}_{k,l})_{k,l = 0, \ldots, K-1}
$ and a  \textit{filled}  
version of it, $\mathbf{\Tilde{F}} = (\mathbf{\Tilde{F}}_{k,l})_{k,l = 0, \ldots, K-1} ,$ and on a 
noise level parameter $\eta \leq 0.5.$
The meta-graph adjacency matrix $\mathbf{F}$ is generated from the given meta-graph structure, called $\mathcal{M}.$  To include an ambient background,
the filled meta-graph adjacency matrix $\mathbf{\Tilde{F}}$ 
replaces every zero in $\mathbf{F}$ that is not part of the imbalance structure by 0.5. The filled meta-graph thus creates a number of {\it ambient nodes} which correspond to entries which are not part of 
$\mathcal{M}$
and thus are not part of a meaningful cluster; this set of {\it ambient nodes} is also called the {\it ambient cluster}. 
First, we provide examples of structures of $\mathbf{F}$
without any ambient nodes, where $\mathbbm{1}$ denotes the indicator function: 

1)~``\textit{cycle}":   
$
    \mathbf{F}_{k,l}=
    (1-\eta) \mathbbm{1}(l=((k+1)\mod K))+
    \eta\mathbbm{1}(l=((k-1)\mod K)) +
    \frac{1}{2}\mathbbm{1}(l=k).
$ 

2)  
``\textit{path}": 
$
    \mathbf{F}_{k,l}=
    (1-\eta) \mathbbm{1}(l=k+1)+
    \eta\mathbbm{1}(l=k-1)+
    \frac{1}{2}\mathbbm{1}(l=k).
$ 

3) ``\textit{complete}":
assign diagonal entries $\frac{1}{2}.$ For each pair $(k,l)$ with $k < l,$ let  $\mathbf{F}_{k,l}$ be $\eta$ and $1-\eta$ with equal probability, then assign $\mathbf{F}_{l,k}=1-\mathbf{F}_{k,l}.$ 

4)  ``\textit{star}", following \cite{elliottanomaly}:
select the center node as $ \omega = \lfloor\frac{K-1}{2}\rfloor$ and set
$
    \mathbf{F}_{k,l}=
    (1-\eta)\mathbbm{1}(k= \omega , l \text{ odd})+
    \eta\mathbbm{1}( k= \omega , l \text{ even})+
    (1-\eta)\mathbbm{1}(l= \omega , k \text{ odd})+
    \eta\mathbbm{1}( l= \omega , l \text{ even}) .
$

When ambient nodes are present, the construction involves two steps, with the first step the same as the above, but for ``\textit{cycle}" meta-graph structure,
$
    \mathbf{F}_{k,l}=
    (1-\eta) \mathbbm{1}(l=((k+1)\mod (K-1)))+
    \eta\mathbbm{1}(l=((k-1)\mod (K-1))) +
    0.5 \, \mathbbm{1}(l=k).
$
The second step is to assign 0 (0.5, resp.) to the last row and the last column of $\mathbf{F}$ ($\mathbf{\Tilde{F}}$, resp.).

A DSBM model, denoted by 
DSBM ($\mathcal{M}, \mathbbm{1}(\text{ambient}), n, K, p, \rho, \eta$), 
is built as follows: 

1)-2) Same as 1)-2) in Sec.~\ref{subsubsec:ssbm}. 

3) For nodes $v_i, v_j\in\mathcal{C}_k$, independently sample an edge from $v_i$ to $v_j$ with probability $ p \cdot \mathbf{\Tilde{F}}_{k,k}$. 

4) For each pair of different clusters $\mathcal{C}_k,\mathcal{C}_l$ with $k\neq l$, for each node $v_i\in\mathcal{C}_k$, and each node $v_j\in\mathcal{C}_l$, independently sample an edge from $v_i$ to $v_j$ with probability $p \cdot \mathbf{\Tilde{F}}_{k,l}$.

\subsubsection{Signed Directed Stochastic Block Model (SDSBM)}
\label{sec: novel sdsbm}
A signed directed stochastic block model (SDSBM)~\cite{he2022msgnn} relies on a meta-graph adjacency matrix $\mathbf{F} = (\mathbf{F}_{k,l})_{k,l = 0, \ldots, C-1}
$, edge sparsity level $p$, the number of nodes $n$, and on a 
sign flip noise level parameter $0 \leq \eta \leq 0.5$. 
An SDSBM model, denoted by 
SDSBM ($\mathbf{F}, n, p, \rho, \eta$),  
is built as follows: 

1)-2) same as 1)-2) in Sec.~\ref{subsubsec:ssbm}. 

3) For nodes $v_i\in\mathcal{C}_k$, and $v_j\in\mathcal{C}_l$, independently sample an edge from $v_i$ to $v_j$ with probability $ p \cdot \lvert\mathbf{F}_{k,l}\rvert$. Give this edge weight $1$ if $F_{k,l}\geq 0$ and weight $-1$ if $F_{k,l}<0$. 

4) Flip the sign of all the edges in the generated graph with sign flip probability $\eta$.
\section{Real-World Data Set Description}
\label{appendix_sec:real_world_description}
\subsection{Directed Unsigned Networks} 
As real-world directed, unsigned networks we include the following data sets.
\begin{itemize}
\item 
\textit{Blog}~\cite{adamic2005political} records  $\lvert \mathcal{E} \rvert=19,024$ directed edges between $n=1,212$ political blogs from the 2004 US presidential election. 

\item 
\textit{Migration}~\cite{perry2003state} reports the number of people that migrated between pairs of counties in the US during 1995-2000. 
It involves $n=3,075$ countries
and $\lvert \mathcal{E} \rvert=721,432$ directed edges after obtaining the largest weakly connected component.
Since the original directed network has a few extremely large entries, to cope with these outliers we 
preprocess the input network by
$
\mathbf{A}_{i,j}=\frac{\mathbf{A}_{i,j}}{\mathbf{A}_{i,j}+\mathbf{A}_{j,i}}\mathbbm{1}(\mathbf{A}_{i,j}>0), \forall i,j\in\{1,\cdots,n\},
$
which follows
the preprocessing of \cite{cucuringu2020hermitian}. 

\item 
\textit{WikiTalk}~\cite{leskovec2010signed} contains all users and discussions from the inception of Wikipedia until Jan. 2008. 
The $n=2,388,953$ nodes in the network represent Wikipedia users and a directed edge from node $v_i$ to node $v_j$ denotes that user $i$  edited at least once a talk page of user $j.$ We extract the largest weakly connected component. 

\item \textit{Telegram}~\cite{bovet2020activity} is a pairwise influence network between $245$ Telegram channels with $8,912$ edges. Labels are generated from the method discussed in~\cite{bovet2020activity}, with a total of four classes. 

\item \textit{Cora-ML}~\cite{bojchevski2017deep} is a popular citation network with node labels based on paper topics with seven classes. In this citation network, nodes represent papers, edges denote
citations of one paper by another, and node features are the bag-of-words representation of papers. The resulting network has 2,995 nodes and 8,416 edges.

\item \textit{CiteSeer} \cite{giles1998citeseer} is a citation data set from an automatic citation indexing system. In this citation network, nodes represent papers, and edges denote
citations of one paper by another. Node features are the bag-of-words representation of papers, and
node labels are determined by the academic topic of a paper. The resulting network has 3,312 nodes and 4,715 edges.

\item \textit{Texas}, \textit{Wisconsin}, and \textit{Cornell} are WebKB data sets extracted from the CMU World Wide Knowledge Base (Web->KB) project\footnote{\url{http://www.cs.cmu.edu/afs/cs.cmu.edu/project/theo-11/www/wwkb/}}. They record hyperlinks between websites at different universities. WebKB is a webpage data set collected from computer science departments of various
universities by Carnegie Mellon University. In these networks, nodes represent web pages, and edges are hyperlinks between them. Node features are the bag-of-words representation of web pages. The web pages are manually classified
into the five categories, student, project, course, staff, and faculty~\cite{pei2020geom}. The resulting networks have 183, 251, and 183 nodes respectively, along with 325, 515, and 298 edges, respectively.

\item \textit{Chameleon} and \textit{Squirrel}~\cite{rozemberczki2019multi} represent links between Wikipedia pages related to chameleons and squirrels. In these networks, nodes denote web pages and edges represent mutual links between them. Node features correspond to several informative nouns in the Wikipedia pages. The nodes are classified by~\cite{pei2020geom} into five categories in terms of the number of the average
monthly traffic of the web page. The resulting networks have 2,277 and 5,201 nodes respectively, along with 36,101 and 222,134 edges, respectively.

\item \textit{WikiCS}~\cite{mernyei2020wiki} is a directed network whose nodes correspond to Computer Science articles, and edges are based on hyperlinks. This network has 10 classes resenting different branches of the field. The resulting network has 11,701 nodes and 297,110 edges.

\item
\textit{Lead-Lag}~\cite{bennett2021detection} contains yearly lead-lag matrices from 269 stocks from 2001 to 2019. Each lead-lag matrix is built from a time series of daily price log returns, as detailed in~\cite{bennett2021detection}. The lead-lag metric for entry $(i,j)$ in the network encodes a measure of the extent to which stock $i$ leads stock $j$, and is obtained by applying a functional that computes the signed normalized area under the curve (auc) of the standard cross-correlation function (ccf). The resulting matrix is skew-symmetric, and entry $(i,j)$ quantifies the extent to which stock $i$ leads or lags stocks $j$, thus leading to a directed network interpretation.
Starting from the skew-symmetric matrix, we further convert negative  entries to zero, so that the resulting directed network can be directly fed into other methods; 
note that this step does not throw away any information, and is pursued only to render the representation of the directed network consistent with the format expected by all methods compared. The average number of edges is 29,159.
\end{itemize}
\subsection{Signed Undirected and Signed Directed Networks}
For signed networks, we provide data loaders for the following data sets.
\begin{itemize}
\item The Sampson monastery data~\cite{sampson1969novitiate} covers 4 social relationships, each of which could be positive or negative. 
We combine these relationships into a network of 25 nodes.
For this data set, we use as node attribute whether or not they attended the minor seminary of "Cloisterville".
As ground truth we take Sampson's division of the novices into four groups: Young Turks, Loyal Opposition, Outcasts, and an interstitial group. The number of positive and negative edges are 148 and 182, respectively.

\item \textit{Rainfall}~\cite{bertolacci2019climate} contains Australian rainfalls pairwise correlations. This data set is based on the analysis over 294 million daily rainfall measurements since 1876, spanning 17,606 sites across continental Australia. The data set is further processed in \cite{he2022sssnet}. The resulting network has 306 nodes, and the number of positive and negative edges are 64,408 and 29,228, respectively.

\item \textit{Fin-YNet}~\cite{SP1500, he2022sssnet} consists of yearly correlation matrices for $n=451$ stocks for 2000-2020 (21 distinct networks), using so-called \textit{market excess returns}; that is, we compute each correlation matrix from overnight (previous close to open) and intraday (open-to-close) price daily returns, from which we subtract the market return of the S\&P500 index. The resulting networks have on average 148,527 positive edges and 54,313 negative edges.

\item \textit{S\&P1500}~\cite{SP1500} considers daily prices for $n=1,193$ stocks, in the S\&P 1500 Index, between 2003 and 2015, and builds correlation matrices also from market excess returns.
The result is a fully-connected weighted network, with stocks as nodes and correlations as edge weights. The resulting network has 1,069,319 positive edges and 353,930 negative edges.

\item \textit{PPI}~\cite{vinayagam2014integrating} is a signed protein-protein interaction (PPI) network. The edge signs represent activation-inhibition relationships. This is a Drosophila melanogaster signed PPI network consisting of 6,125 signed PPIs connecting 3,352 proteins that can be used to identify positive and negative regulators of signaling pathways and protein complexes. The data set is further processed to keep the largest connected component. The resulting network has 3,058 nodes, 7,996 positive edges, and 3,864 negative edges.

\item \textit{Wiki-Rfa}~\cite{west2014exploiting} is a signed network describing voting information for electing Wikipedia managers. Positive edges represent supporting votes, while negative edges represent opposing votes. The data set is further processed in \cite{he2022sssnet} to keep only the largest weakly connected component and to remove nodes with very low degrees. The resulting network has 7,634 nodes, 135,753 positive edges, and 37,579 negative edges.

\item \textit{BitCoin-Alpha} and \textit{BitCoin-OTC}~\cite{kumar2016edge} describe bitcoin trading. As a cryptocurrency, Bitcoin is used to trade anonymously over the web, whose counterparty risk~\cite{moore2013beware} has led to the emergence of several exchanges where Bitcoin users rate the level of trust they have in other users. Two such exchanges are OTC (for short) and Alpha (for short). Both exchanges enable users to rate others on a scale of -10 to 10 (excluding zero), where a rating of -10 should be given to fraudsters while 10 means to trust the person as trusting oneself. The rating values in between have intermediate meanings. The resulting networks have 3,783 and 5,881 nodes respectively. \textit{BitCoin-Alpha} has 22,650 positive edges and 1,536 negative edges, while \textit{BitCoin-OTC} has 32,029 positive edges and 3,563 negative edges. 

\item \textit{Slashdot}~\cite{slashdot} relates to a technology-related news website. This network contains friend/foe links between the users of Slashdot. The resulting network has 82,140 nodes, 380,933 positive edges, and 119,548 negative edges.

\item \textit{Epinions}~\cite{epinions} describes trust-distrust consumer reviews on \url{epinions.com}. \url{epinions.com} is a website in which users can write reviews about products and assign them a rating. This website also allows the users to express their Web of
Trust, i.e. ``reviewers whose reviews and ratings they
have consistently found to be valuable" and their Block
list, i.e. a list of authors whose reviews they find consistently offensive, inaccurate, or in general not valuable. Inserting a user in the Web of Trust is the same as issuing a trust statement which is recorded as a positive edge between the user and the reviewer, while inserting them in the Block List means issuing a distrust statement which is recorded as a negative edge. The resulting network has 131,580 nodes, 589,888 positive edges, and 121,322 negative edges. 

\item \textbf{Fi}nancial \textbf{L}ead-\textbf{L}ag (FiLL) data sets. For each year in the data set, \cite{he2022msgnn} builds a signed directed graph (\textit{FiLL-pvCLCL}) based on the price return of 444 stocks at market close times on consecutive days. \cite{he2022msgnn} also builds another graph (\textit{FiLL-OPCL}), based on the price return of 430 stocks from market open to close. The lead-lag metric that is captured by the entry $\mathbf{A}_{i,j}$ in each network encodes a measure that quantifies the extent to which stock $v_i$ leads stock $v_j$, and is obtained by computing the linear regression coefficient when regressing the time series (of length 245) of daily returns of stock $v_i$ against the lag-one version of the time series (of length 245) of the daily returns of stock $v_j$. Specifically, \cite{he2022msgnn} uses the beta coefficient of the corresponding simple linear regression as the one-day lead-lag metric. The resulting matrix is asymmetric and signed, rendering it amenable to a signed directed network interpretation. The data matrices stored in PyGSD are dense but there is a sparsification parameter provided for the data loader; \cite{he2022msgnn} uses a parameter value of 0.2. The resulting annual graphs \textit{FiLL-OPCL} have on average 84,467 positive edges and 100,013 negative edges, while the resulting annual graphs \textit{FiLL-pvCLCL} have on average 84,677 positive edges and 112,015 negative edges.
\end{itemize}
\section{Implementation Details}
\label{appendix:implementation}
The original signed GNN implementations for SGCN\footnote{https://github.com/benedekrozemberczki/SGCN}, SNEA\footnote{https://github.com/liyu1990/snea}, SiGAT, and SDGNN\footnote{https://github.com/huangjunjie-cs/SiGAT} are based on the code of GraphSAGE\footnote{https://github.com/williamleif/graphsage-simple}.
In our implementations, we inherit the class of MessagePassing in PyG and add new GraphConv operators (e.g., SNEAConv) for signed networks, because this combination is more efficient and unified than the original implementations, which is conducive to the generalization of the code. Tables~\ref{tab:link_sign_runtime}, \ref{tab:directed_gcn_time} and \ref{tab:runtime_link_pred} report average runtime of our implemented methods on various tasks, showing that we have efficient implementations, with at least as short a runtime as the original implementations. For the other models, the original implementations are based on PyG, so there is not much an efficiency change compared to them. Note that as SSSNET, DIGRAC, and MagNet are not originally written with the MessagePassing class in PyG, we have transformed the model classes accordingly. We observe comparable (SGCN, SSSNET, DIGRAC, and MSGNN) or even better (SiGAT, SDGNN, SNEA, and MagNet) runtime efficiency for these GNN implementations. For the other GNN implementations (DGCN, DiGCN, DiGCNIB, and DiGCL), we clean up the original codes, add documentation for them, unify some functions, split up some auxiliary layers, add unit tests and provide examples.
\paragraph{Node Clustering.}
We use the same settings as in \cite{he2022sssnet}, \cite{he2022digrac}, and \cite{he2022msgnn} respectively for the node clustering task on signed undirected, unsigned directed, and signed directed graphs, respectively. For all the experiments, we use Adam~\cite{kingma_adam_2014} as our optimizer with learning rate 0.01 and employ $\ell_2$ regularization with weight decay parameter $5\cdot 10^{-4}$ to avoid overfitting. For SSSNET \cite{he2022sssnet} and DIGRAC~\cite{he2022digrac}, we use hidden size 16, 2 hops, and $\tau=0.5$. For MSGNN~\cite{he2022msgnn}, we take a hidden size of 16 and use two layers of convolution. For all synthetic models, we train the GNNs with a maximum of 1000 epochs, and stop training when no gain in validation performance is achieved for 200 epochs (early-stopping). Note that we adapt SSSNET and DIGRAC from matrix multiplication implementations to MessagePassing (from PyG) implementations for their models, and we observe comparable runtime efficiency for our implementations and the original implementations. Specially, for the node clustering results reported in Figures~\ref{fig:DIGRAC} and \ref{fig:SSSNET}, both implementations of DIGRAC consume roughly 0.05 seconds for each epoch, while both implementations of SSSNET consume roughly 0.4 seconds per epoch. Experiments were conducted on a compute node with 8 Nvidia Tesla T4, 96 Intel Xeon Platinum 8259CL CPUs @ 2.50GHz and $378$GB RAM.

For all node clustering tasks, we set 10\% of all nodes from each cluster as test nodes, 10\% as validation nodes to select the model, and the remaining 80\% as training nodes. For signed networks, we are in a semi-supervised setting, where 10\% of the training nodes are set to be seed nodes. Note that label supervision is not provided for all training nodes but only for seed nodes in SSSNET, DIGRAC, and MSGNN, while the subgraph induced by the set of training nodes is the graph to which the self-supervised loss function is applied. The other GNNs compared here (DiGCL, DiGCN, DiGCNIB, MagNet and DGCN) are trained in a fully-supervised manner, meaning that labels for all training nodes are used during training, and also the self-supervised loss function is not applied to them, see \cite{he2022digrac} for more details. For the DSBM experiments, we generate 5 DSBM networks under each parameter setting and use 10 different data splits for each network, then average over the 50 runs. For the remaining synthetic models, we first generate five different networks, each with two different data splits, then conduct experiments on them and report average performance over these 10 runs. 
The meta-graph adjacency matrices for the SDSBM models are defined as
$$\mathbf{F}_1(\gamma)=\begin{bmatrix}
0.5 & \gamma & -\gamma\\
1-\gamma & 0.5 & -0.5\\
-1+\gamma&-0.5&0.5
\end{bmatrix},\mathbf{F}_2(\gamma)=\begin{bmatrix}
0.5 & \gamma & -\gamma&-\gamma\\
1-\gamma & 0.5 &-0.5& -\gamma\\
-1+\gamma&-0.5&0.5&-\gamma\\
-1+\gamma&-1+\gamma&-1+\gamma&0.5
\end{bmatrix}.$$ 

In Sec.~\ref{sec:methods} also some non-GNN methods are mentioned, and they are included in the experiments in Sec.~\ref{experiments}. In detail,

Herm and Herm\_sym are two variants from \cite{cucuringu2020hermitian}, L and L\_sym are two Signed Laplacian variants from~\cite{kunegis2010spectral}, sns and dns are two Laplacian variations from~\cite{mercado2016clustering}, BNC and BRC are abbreviations for Balanced Normalized Cut and Balanced Ratio Cut from~\cite{Chiang},  and finally SPONGE and SPONGE\_sym are two variants from~\cite{SPONGE_AISTATS_2019}.
\paragraph{Link Sign Prediction on Signed Graphs.}
Link Sign Prediction on Signed Graphs is the task to predict the sign of given edges in a network, which is a binary classification task.
Considering the label imbalance, F1 and AUC are used to evaluate the performance.
Past signed GNNs on this task adopt a two-stage process, first using signed network embedding methods to learn representations from the train set, and then using a downstream classifier to predict the signs of edges in the test set.
However, the setting is not conducive to searching hyper-parameters.
In this paper, we split 80\% edges for training, 10\% edges for validation, and the remaining edges constitute the test set.
We run such train/validation/test splits five times to  report the mean and standard deviation over the five runs.
The downstream classifier is a logistic regression classifier with 80\% train edges.
The dimension number of node representations $\mathbf{Z}$ is set to 20 for all the methods, which is widely used in related works. 
We use Adam optimizer to optimize all the models (The learning rate is set to 0.01 and the weight decay is set to 1e-3).
We run 500 epochs (200 epochs as warming up) with 10 epoch evaluations on the AUC metric on the test set and adopt early stopping with the patience of 10 epochs to prevent overfitting.
For other parameters, we follow the authors' suggestions and the released codes. Experiments were conducted on a compute node with an NVIDIA Tesla V100S GPU (32GB) and Intel(R) Xeon(R) Silver 4310 CPUs (250GB). Table~\ref{tab:link_sign_runtime} reports the average runtime per epoch for both our implementations and the original ones, where we observe a huge improvement in runtime efficiency over the original implementations for SiGAT, SNEA and SDGNN. For SGCN, the original and our implementations consume comparable runtime, and hence we omit the runtime report for the original implementation here. 

\begin{table}[ht]
\caption{Link sign prediction average runtime per epoch (in seconds) for signed networks, where "ours" denotes our implementation, and "original" denotes the original implementation. The fastest is marked in \textbf{bold}.}
\centering
\small 
\setlength\tabcolsep{3pt}
 \resizebox{0.8\linewidth}{!}{
\begin{tabular}{cccccc}
\toprule
Method & Bitcoin-Alpha & Bitcoin-OTC & Wikirfa & Slashdot & Epinions\\ \midrule
SGCN(ours)            & \textbf{0.08} & \textbf{0.08}  & 0.26  & 0.80   & 1.27                               \\
SiGAT(ours)           & 0.10 & 0.11& \textbf{0.11}    & \textbf{0.14}   & \textbf{0.19}                               \\
SiGAT(original) & 2.93 & 4.71  & 51.83 & 121.56 & 236.53                             \\
SDGNN(ours)           & \textbf{0.08} & 0.07  & 0.12  & 0.25   & 0.41                               \\
SDGNN(original) & 7.30 & 12.06 & 93.21 & 359.57 & \multicolumn{1}{l}{out of memory}  \\
SNEA(ours)            & 0.09 & 0.16  & 0.38  & 0.91   & 1.41                               \\
SNEA(original)  & 0.92 & 2.16  & 10.11 & 220.15 & 431.42  \\
\bottomrule
\end{tabular}
}
\label{tab:link_sign_runtime}
\end{table}

\paragraph{Node Classification and Edge-Level Tasks on Directed Unsigned Graphs.} We adopt the settings in \cite{zhang2021magnet} for MagNet, DGCN, DiGCN, and DiGCNIB, for directed network node classification and edge-level tasks. For DiGCL results, we use the settings in \cite{tong2021directed} for Cora\_ML and CiteSeer for node classification and search parameters for the other data sets. For link prediction tasks, we follow the settings from \cite{zhang2021magnet}, training all GNNs for each link prediction task for 3000 epochs and set the early stopping as 500. We tune the number of filters in $[16, 32, 64]$ for MagNet, DGCN, DiGCN, and search the number of filters of DiGCNIB in $[6, 11, 21]$. The coefficient $\alpha$ of DiGCN and DiGCNIB is searched in $[0.05, 0.1, 0.15, 0.2]$. The $q$ parameter in MagNet is tuned in $[0.05, 0.1, 0.15, 0.2, 0.25]$. Table~\ref{tab:directed_gcn_time} reports the total runtime for the node classification and link prediction tasks on directed unsigned networks with $2$ layers of different directed graph convolution, training with $10$ folds $\times$ $500$ epochs per fold. $32$ filters are used in DGCN, DiGCN, DiGCNIB and MagNet. For DiGCL, $128$ filters are used for GCNs and $64$ filters are used for the projection. As shown in the table, our implementations are efficient. The running time of directed unsigned graphs is evaluated on a compute node with 16 threads of i9-11900KF CPU, RTX3090 GPU, and 32GB memory. We adopted the original implementation for directed graph convolution layers except for MagNet. We developed MagNet with MessagePassing from PyG in our package. The comparison of the MagNet runtimes validates our efficiency improvement.

\begin{table}[htp!]
\caption{Runtime (seconds) comparison on link tasks. The fastest is marked in \textbf{bold}. MagNet (original) denotes the original implementation and MagNet (ours) denotes the implementation of our package.}
\begin{center}
\resizebox{1.0\linewidth}{!}{\begin{tabular}{ccccccccc}
\toprule
Task&Method&Cornell&Texas&Wisconsin&CoraML&CiteSeer&Telegram \\
\midrule 
\multirow{5}{*}{Node classification}&
DGCN&16&17&17&43&43&17\\
&DiGCN&\textbf{9}&\textbf{9}&\textbf{9}&\textbf{32}&\textbf{37}&\textbf{10}\\
&DiGCNIB&19&19&19&42&46&20\\
&DiGCL&28&29&30&63&57&37\\
&MagNet (original)&73&74&81&215&173&54\\
&MagNet (ours)&21&21&21&96&94&20\\
\midrule
\multirow{5}{*}{Direction prediction}
&DGCN&30 &30 &31 &56 &55 &34 \\
&DiGCN&\textbf{10} &\textbf{10} &\textbf{10} &\textbf{31} &\textbf{34} &\textbf{11} \\
&DiGCNIB&19 &19 &19 &41 &42 &21 \\
&Magnet(original)&53 &52 &52 &61 &57 &56 \\
&MagNet(ours)&36 &36 &36 &40 &39 &43 \\
\midrule
\multirow{5}{*}{Existence link prediction}
&DGCN&29 &29 &30 &58 &56 &35 \\
&DiGCN&\textbf{10} &\textbf{9} &\textbf{10} &\textbf{31} &\textbf{34} &\textbf{13} \\
&DiGCNIB&19 &20 &19 &40 &43 &22 \\
&Magnet(original)&51 &52 &53 &63 &59 &60 \\
&MagNet(ours)&36 &37 &37 &42 &40 &43 \\
\midrule
\multirow{5}{*}{Three classes link prediction}
&DGCN&30 &30 &31 &59 &55 &36 \\
&DiGCN&\textbf{10} &\textbf{10} &\textbf{10} &\textbf{33} &\textbf{34} &\textbf{13} \\
&DiGCNIB&19 &19 &19 &42 &44 &22 \\
&Magnet(original)&52 &52 &52 &61 &57 &59 \\
&MagNet(ours)&36 &37 &37 &41 &39 &41 \\
\bottomrule
\end{tabular}}
\end{center}
\label{tab:directed_gcn_time}
\end{table}
\paragraph{Link Prediction on Signed Directed Graphs.} We follow the settings from \cite{he2022msgnn}, training all GNNs for each link prediction task for 300 epochs. We use the proposed loss functions from their original papers for SGCN \cite{Derr}, SNEA \cite{li2020learning}, SiGAT \cite{huang2019signed}, and SDGNN \cite{huang2021sdgnn}, and we use the cross-entropy loss for SSSNET \cite{he2022sssnet} and MSGNN~\cite{he2022msgnn}. For all link prediction experiments in Table~\ref{tab:link_sign_direction_prediction_performance}, we sample 20\% edges as test edges, and use the rest of the edges for training. Five splits were generated randomly for each input graph. We calculate the in- and out-degrees based on both signs from the observed input graph (removing test edges) to obtain a four-dimensional feature vector for each node for training SSSNET \cite{he2022sssnet}, and MSGNN~\cite{he2022msgnn}, and we use the default settings from relative papers for SGCN \cite{Derr}, SNEA \cite{li2020learning}, SiGAT \cite{huang2019signed}, and SDGNN \cite{huang2021sdgnn}. Note that the ``avg." results for \textit{FiLL} first average the accuracy values across all individual networks (a total of 42 networks), then report the mean and standard deviation over the five runs. Table~\ref{tab:runtime_link_pred} reports average runtimes for the experiments. Experiments were conducted on a compute node with 8 Nvidia Tesla T4, 96 Intel Xeon Platinum 8259CL CPUs @ 2.50GHz and $378$GB RAM.

\begin{table*}
\caption{Runtime (seconds) comparison on link tasks. The fastest is marked in \textbf{bold}.}
  \label{tab:runtime_link_pred}
  \centering
  \small
  \resizebox{\linewidth}{!}{\begin{tabular}{c c c c c  ccccc}
    \toprule
    Data Set& Task&GCN&SGCN&SDGNN&SiGAT&SNEA&SSSNET&MSGNN \\
    \midrule
\multirow{4}{*}{BitCoin-Alpha}&SP&\textbf{23}&352&124&277&438&59&29\\
&DP&\textbf{24}&328&196&432&498&78&37\\
&3C&\textbf{32}&403&150&288&446&77&37\\
&4C&\textbf{30}&385&133&293&471&57&36\\
&5C&\textbf{31}&350&373&468&570&82&37\\
\midrule
\multirow{4}{*}{BitCoin-OTC}&SP&\textbf{27}&340&140&397&584&68&30\\
&DP&\textbf{26}&471&243&426&941&80&38\\
&3C&\textbf{37}&292&252&502&551&92&\textbf{37}\\
&4C&\textbf{31}&347&143&487&607&68&37\\
&5C&\textbf{37}&460&507&500&959&86&38\\
\midrule
\multirow{4}{*}{Slashdot}&SP&\textbf{167}&4218&3282&1159&5792&342&227\\
&DP&\textbf{171}&4231&3129&1200&5773&311&222\\
&3C&\textbf{208}&3686&6517&1117&6628&263&322\\
&4C&\textbf{133}&4038&5296&948&7349&202&232\\
&5C&\textbf{212}&4269&7394&904&8246&424&327\\
\midrule
\multirow{4}{*}{Epinions}&SP&\textbf{272}&6436&4725&2527&8734&300&370\\
&DP&\textbf{273}&6437&4605&2381&8662&404&369\\
&3C&\textbf{322}&6555&8746&2779&10536&471&510\\
&4C&\textbf{214}&6466&6923&2483&10380&272&384\\
&5C&\textbf{329}&7974&9310&2719&11780&460&517\\
\midrule
\multirow{4}{*}{FiLL (avg.)}&SP&\textbf{29}&591&320&367&617&61&32\\
&DP&\textbf{30}&387&316&363&386&53&36\\
&3C&\textbf{36}&542&471&298&657&79&43\\
&4C&\textbf{27}&608&384&343&642&56&35\\
&5C&\textbf{38}&318&534&266&521&63&44\\
\bottomrule
\end{tabular}}
\end{table*}

\end{document}